\pdfoutput=1

\documentclass[11pt, twocolumn]{article}

\usepackage[preprint]{acl}

\usepackage{times}
\usepackage{latexsym}
\usepackage[most]{tcolorbox}

\usepackage[T1]{fontenc}

\usepackage[utf8]{inputenc}

\usepackage{microtype}
\usepackage{fancyvrb}
\usepackage{inconsolata}

\usepackage{multirow}           
\usepackage{booktabs}           
\usepackage{arydshln}
\usepackage{graphicx}           
\usepackage{float}              
\usepackage{bm}                 
\usepackage[tikz]{bclogo}       
\usepackage{amsmath}            
\usepackage{amsfonts}           
\usepackage{caption}
\usepackage{subcaption}         
\usepackage{pifont}             
\usepackage{enumitem}           

\newcommand{\Exclusivity}{Exclusivity}
\newcommand{\Coverage}{Coverage}
\newcommand{\Balance}{Balance}
\newcommand{\Usability}{Usability}

\title{Revisiting Classification Taxonomy for Grammatical Errors}

\author{
    \textbf{Deqing Zou}\textsuperscript{1}\thanks{Equal contribution.},
    \textbf{Jingheng Ye}\textsuperscript{1}\footnotemark[1],
    \textbf{Yulu Liu}\textsuperscript{2}, 
    \textbf{Yu Wu}\textsuperscript{1}, 
    \textbf{Zishan Xu}\textsuperscript{1}, \\
    \textbf{Yinghui Li}\textsuperscript{1}, 
    \textbf{Hai-Tao Zheng}\textsuperscript{1,3}\thanks{Corresponding author: Hai-Tao Zheng. \texttt{(Email: zheng.haitao@sz.tsinghua.edu.cn)}}, 
    \textbf{Bingxu An}\textsuperscript{4},
    \textbf{Zhao Wei}\textsuperscript{4}, 
    \textbf{Yong Xu}\textsuperscript{4} \\
    \textsuperscript{1}Tsinghua University, \\
    \textsuperscript{2}University of Electronic Science and Technology of China, \\
    \textsuperscript{3}Peng Cheng Laboratory,
    \textsuperscript{4}Tencent \\
    \texttt{\{zdq23,yejh22\}@mails.tsinghua.edu.cn}
}

\begin{document}
\maketitle

\begin{abstract}
Grammatical error classification plays a crucial role in language learning systems, but existing classification taxonomies often lack rigorous validation, leading to inconsistencies and unreliable feedback. In this paper, we revisit previous classification taxonomies for grammatical errors by introducing a systematic and qualitative evaluation framework. Our approach examines four aspects of a taxonomy, i.e., exclusivity, coverage, balance, and usability. Then, we construct a high-quality grammatical error classification dataset annotated with multiple classification taxonomies and evaluate them grounding on our proposed evaluation framework. Our experiments reveal the drawbacks of existing taxonomies. Our contributions aim to improve the precision and effectiveness of error analysis, providing more understandable and actionable feedback for language learners.\footnote{All the codes and data will be released after the review.}
\end{abstract}

\section{Introduction}

Errors are an inevitable aspect of language acquisition, serving as critical indicators of learners' linguistic development and providing valuable insights for educators and intelligent language learning systems \citep{l10,l11,l12,liu2022we,li2022past,li2023towards,li2024llms}. In Error Analysis (EA), the systematic identification, categorization, and interpretation of learner errors play a pivotal role in improving personalized instruction, generating automated feedback, and enabling effective language assessment \citep{l13,l14,language_two,li2022learning,li2023effectiveness}. Central to this process is the use of grammatical error classification taxonomies~\citep{ma2022linguistic,excgec,enhancing}, which organize learner errors based on linguistic or cognitive principles. These taxonomies have been widely adopted in applications such as grammatical error correction (GEC)~\cite{li2024rethinkingroleslargelanguage,ye-etal-2023-mixedit,ye-etal-2023-system,li2025correct} and automated essay scoring (AES), significantly enhancing error detection \cite{huang-etal-2023-frustratingly,zhang2023contextual}, correction~\cite{ye2022focus}, and feedback generation \citep{l15,l16}.

A well-designed grammatical error classification taxonomy allows language learners to better understand the nature and causes of their errors, facilitating targeted improvements in their linguistic competence \cite{ye2025position}. However, existing taxonomies are often developed based on empirical assumptions or ad-hoc practices, without rigorous validation \citep{TEGA, errant}. This lack of systematic evaluation has led to \textit{issues} such as overlapping categories, insufficient coverage of error types, and limited applicability in real-world educational contexts.

To address these challenges, this paper revisits grammatical error classification taxonomies by systematically assessing their quality and utility. Specifically, we introduce a multi-metric evaluation framework that examines four key dimensions of a taxonomy: \ding{182} \textit{\Exclusivity} ensures that error categories are mutually exclusive, with clearly defined boundaries to minimize overlap and ambiguity;
\ding{183} \textit{\Coverage} evaluates the extent to which the taxonomy captures both common and rare error types, ensuring a comprehensive representation of learner errors. \ding{184} \textit{\Balance} measures the taxonomy’s ability to balance attention between frequent and infrequent error types, avoiding overemphasis on a narrow subset of errors; \ding{185} \textit{\Usability} assesses the clarity and practical applicability of the taxonomy.

To validate our evaluation framework, we construct a high-quality grammatical error dataset annotated with multiple classification taxonomies. This dataset is created through a collaborative annotation approach that leverages large language models (LLMs) and human annotators, ensuring scalability and annotation reliability. Using this dataset, we systematically evaluate four widely-used error classification taxonomies: \textbf{POL73} \citep{linguistic_one}, \textbf{TUC74} \citep{gooficon}, \textbf{BRY17} \citep{errant}, and \textbf{FEI23} \citep{enhancing}. Through performance comparisons and annotator agreement experiments, we evaluate the rationality of these taxonomies. An ablation study on error type merging further reveals that classification taxonomies should not rely solely on empirical intuition but require systematic validation. These findings underscore the necessity of a rigorous evaluation for grammatical error taxonomy. In summary, our contributions are as follows:

\begin{itemize}[leftmargin=*]
\item[$\bullet$] We propose a novel multi-metric evaluation framework for grammatical error classification taxonomies, incorporating dimensions of exclusivity, coverage, balance, and usability.

\item[$\bullet$] We construct a high-quality grammatical error dataset annotated with multiple taxonomies, leveraging a collaborative annotation process involving LLMs and human experts.

\item[$\bullet$] We conduct a comprehensive evaluation of four widely used taxonomies, providing insights into their strengths, limitations, and practical implications for error analysis in language learning.
\end{itemize}

\section{Related Work}
Existing studies propose various error classification methods based on different perspectives, such as linguistic structure, cognitive processes, and communicative impact. \citet{language_two} categorize language errors into four primary taxonomies: 
(1) \textit{Linguistic Category Taxonomy} \citep{linguistic_one,gooficon,linguistic_three} classifies errors based on language components or linguistic constituents;
(2) \textit{Surface Strategy Taxonomy} \citep{surface_one,surface_two,surface_three} focuses on structural modifications made by learners;
(3) \textit{Comparative Taxonomy} \citep{comparative_one,comparative_child,comparative_adult} classifies errors by comparing L2 learner errors to L1 acquisition errors or native language structures;
(4) \textit{Communicative Effect Taxonomy} \citep{communicative_one,communicative_two} classify errors based on their communicative impact. In addition to the above systematic classification taxonomies, other studies \citep{other_two,NUCLE,other_four,enhancing} have introduced other taxonomies, such as the simplification strategy taxonomy \citep{other_one} and the rule-based annotation toolkit \citep{errant,ye-etal-2023-cleme}.

\section{Methodology}
Given a dataset $D$ of English learner texts, our goal is to systematically and quantitatively evaluate the rationality of $n$ error classification taxonomies $F={\{F_1, F_2,...,F_n\}}$. Each taxonomy $F_i$ consists of $m$ predefined error types, denoted as $F_i=\{ET_1, ET_2,...,ET_m\}$, where $ET_j$ represents the $j$-th error type in the taxonomy $F_i$. All notations are detailed in the Appendix~\ref{appendix notation table}.

We evaluate the rationality of each taxonomy along four dimensions: \textbf{Exclusivity}, \textbf{Coverage}, \textbf{Balance}, and \textbf{Usability}. Exclusivity and Usability are based on the generated results by the LLM, while Balance and Coverage are computed obtained directly from the manually annotated dataset through statistical analysis.

\subsection{Exclusivity}
The error types in the taxonomy should be mutually exclusive, meaning that each error instance belongs to a single distinct category. Overlapping error types introduce instability and inconsistencies in error analysis, reducing the reliability of classification results. If a model frequently assigns high confidence to multiple categories for the same error, it suggests that the taxonomy lacks clear boundaries between certain error types. To quantify this issue, we assess exclusivity by analyzing the confidence scores of an LLM’s predictions. Confidence estimation plays a crucial role in this evaluation, we incorporate three established methods \citep{confidence} for improved reliability (details are in Appendix \ref{appendix confidence}). Specifically, a sample is considered classified under an error type if its confidence score exceeds the predefined threshold $\tau$. We define the $\operatorname{Overlap}$ to quantify the degree of category overlap for a sample $x$ as follows:
\begin{equation}\begin{small}\begin{aligned}
\operatorname{Overlap}(x)=|\{\hat{Y}_i^{(x)} \mid C_i^{(x)} > \tau\}|,
\end{aligned}\end{small}\end{equation}
\noindent where $\hat{Y}_i^{(x)}$ denotes the $i$-th predicted error type for a given sample $x$, and $C_i^{(x)}$ represents its confidence score. A higher $\operatorname{Overlap}(x)$ indicates that multiple error types are assigned to the same sample, suggesting a violation of exclusivity.

The \textit{Exclusivity Score} is computed as the average instance-level exclusivity over the dataset $D$:
\begin{equation}\begin{small}\begin{aligned}
& \operatorname{Exclusivity}(F)= \\
& ~~~~~~~~\frac{1}{|D|}\sum_{x\in D}\begin{cases}
    1- \frac{\operatorname{Overlap}(x) - 1}{k - 1}, & \text{ if } \operatorname{Overlap}(x) > 0, \\
    0, & \text{ if } \operatorname{Overlap}(x) > 0,
\end{cases}
\end{aligned}\end{small}\end{equation}
\noindent where $k$ represents the selection parameter in the Top-K Prompting Strategy, detailed in Appendix \ref{appendix confidence}. $\operatorname{Exclusivity}$ indicates whether the classification taxonomy maintains clear distinctions between error types. A lower score suggests significant overlap between categories, indicating poorly defined boundaries, whereas a higher score implies a more distinct and reliable classification system.

\subsection{Coverage}
Coverage measures the extent to which a taxonomy accounts for errors in the dataset $D$. Let $|U|$ be the number of errors labeled with at least one defined category (as opposed to ``Other''), and $|D|$ be the total number of error instances. We define \textit{Coverage score} as follows:
\begin{equation}\begin{small}\begin{aligned}
\operatorname{Coverage}(F)=\frac{|U|}{|D|}.
\end{aligned}\end{small}\end{equation}

A higher $\operatorname{Coverage}$ indicates that the taxonomy captures a greater proportion of actual errors, demonstrating superior completeness in covering the range of error types.

\subsection{Balance}
Error classification taxonomy should maintain a balanced distribution of error types, avoiding excessive concentration on a few dominant categories while ensuring sufficient representation of less frequent errors. An imbalanced taxonomy may exhibit a long-tail effect \citep{lt1,lt2}, where frequent error types overshadow less common yet pedagogically or computationally significant ones, leading to biased analysis. To assess the balance of a classification taxonomy, we introduce the \textit{Balance Score}. This metric quantifies the evenness of error type distribution using entropy-based uniformity. Given an error type $ET_i$ with proportion computed as:
\begin{equation}\begin{small}\begin{aligned}
    P_i=\frac{|ET_i|}{\sum_{j=1}^m|ET_j|},
\end{aligned}\end{small}\end{equation}
\noindent its entropy contribution is $-P_i \log P_i$ if $|ET_i|>0$, otherwise 0. The final score is normalized by $\log(m)$, where $m$ is the total number of error types:
\begin{equation}\begin{small}\begin{aligned}
\operatorname{Balance}(F)=\frac{\sum_{i=1}^m -P_i \log P_i}{\log (m)}.
\end{aligned}\end{small}\end{equation}

A greater $\operatorname{Balance}$ value signifies a more uniform distribution, attenuating the long-tail effect and guaranteeing its applicability within educational and computational domains.

\begin{table*}[htb]
    \centering
    \resizebox{\textwidth}{!}{
    \begin{tabular}{lcccc cccc cccc cccc} 
        \toprule
        \multirow{2}{*}{\textbf{Model}} & \multicolumn{4}{c}{\textbf{Exclusivity ($\tau =0.7)$} $\uparrow$} & \multicolumn{4}{c}{\textbf{Coverage} $\uparrow$} & \multicolumn{4}{c}{\textbf{Balance} $\uparrow$} & \multicolumn{4}{c}{\textbf{Usability (Macro F1 / Micro F1)} $\uparrow$} \\
        \cmidrule(lr){2-5} \cmidrule(lr){6-9} \cmidrule(lr){10-13} \cmidrule(lr){14-17}
        & \textbf{POL73} & \textbf{TUC74} & \textbf{BRY17} & \textbf{FEI23} & \textbf{POL73} & \textbf{TUC74} & \textbf{BRY17} & \textbf{FEI23} & \textbf{POL73} & \textbf{TUC74} & \textbf{BRY17} & \textbf{FEI23}& \textbf{POL73} & \textbf{TUC74} & \textbf{BRY17} & \textbf{FEI23} \\
        \midrule
        Meta-Llama-3-8B-Instruct & 0.477 & 0.772 & \textbf{0.880} & 0.858 & \multirow{4}{*}{0.698} & \multirow{4}{*}{0.160} & \multirow{4}{*}{\textbf{0.979}} & \multirow{4}{*}{0.924} & \multirow{4}{*}{0.687} & \multirow{4}{*}{0.210} & \multirow{4}{*}{0.829} & \multirow{4}{*}{\textbf{0.878}} & 0.026 / 0.109 & 0.005 / 0.006 & 0.047 / 0.156 & \textbf{0.112} / \textbf{0.357} \\
        Mistral-7B-Instruct-v0.2 & 0.317 & 0.095 & 0.495 & \textbf{0.840} &  &  &  &  &  &  &  &  & 0.026 / 0.123 & 0.001 / 0.004 & 0.015 / 0.113 & \textbf{0.085} / \textbf{0.310}\\
        Claude-3-Haiku & 0.609& 0.596 & 0.742 & \textbf{0.815} &  &  &  &  & &  &  &  & 0.101 / 0.218 & 0.023 / 0.055 & \textbf{0.326} / 0.542 & 0.290 / \textbf{0.620} \\
        ChatGPT-4o & 0.842 & 0.703 & \textbf{0.921} & 0.877 &  &  &  &  &  &  &  &  & 0.301 / 0.478 & 0.061 / 0.099 & 0.610 / \textbf{0.760} & \textbf{0.631} / 0.743 \\
        \midrule
        Average & 0.562 & 0.542 & 0.759 & \textbf{0.848} & 0.698 & 0.160 & \textbf{0.979} & 0.924 & 0.687 & 0.210 & 0.829 & \textbf{0.878} & 0.114 / 0.232 & 0.023 / 0.041 & 0.250 / 0.393 & \textbf{0.280} / \textbf{0.508} \\
        \bottomrule
    \end{tabular}}
    \caption{Performance comparison of classification taxonomies. Higher values indicate better performance in all metrics (Exclusivity, Coverage, Balance, and Usability). The computation of Exclusivity and Usability relies on specific LLMs. Exclusivity is calculated using a confidence threshold of $\tau = 0.7$.}
    \label{tab1}
\end{table*}

\subsection{Usability}
We argue that a taxonomy with great usability should be understandable for humans and models. So we quantify usability from \textit{model effectiveness} and \textit{human annotation agreement}. Model effectiveness means that LLMs can produce reliable predictions based on a classification taxonomy, thereby enhancing the validity of subsequent error analyses. To quantify this, we evaluate classification performance using \textit{Macro F1} and \textit{Micro F1} scores:
\begin{equation}\begin{small}\begin{aligned} 
\operatorname{Macro\_F1} &= \frac{1}{m}\sum_{i=1}^m \frac{2 \cdot P(ET_i)\cdot R(ET_i)}{P(ET_i)+R(ET_i)},
\end{aligned}\end{small}\end{equation}
\begin{equation}\begin{small}\begin{aligned}
\operatorname{Micro\_F1} &= \frac{2 \cdot P(D)\cdot R(D)}{P(D)+R(D)},
\end{aligned}\end{small}\end{equation}
\noindent where $P$ and $R$ denote precision and recall, respectively. A high Macro F1 indicates the taxonomy supports stable model performance across both frequent and infrequent error types, mitigating classification bias. A high Micro F1 suggests its robustness in large-scale error detection. The combination of high Macro and Micro F1 scores demonstrates that the taxonomy maintains both category-level consistency and large-scale applicability, ensuring the reliability of error analysis.

For human annotators, the taxonomy should be intuitive and easy to apply, minimizing ambiguity in error categorization. Therefore, we measure inter-annotator consistency in Section \ref{section annotator agreement}.

\section{Experiments}
\subsection{Experimental Setup}

\paragraph{Dataset \& Models}
We use the Cambridge English Write \& Improve (W\&I) and LOCNESS corpus \citep{w&i}, re-annotated with ERRANT \citep{errant}. To reduce the uncertainty introduced by multiple grammatical errors in a sentence, we preprocess the dataset by decomposing sentences with multiple errors into single-error instances \citep{enhancing}, as described in Appendix \ref{appendix extract single}. We then select 487 instances for further analysis. And we adopt an LLM \& human collaborative annotation approach (Appendix \ref{appendix annotation}). A detailed description of the dataset is provided in Appendix~\ref{appendix dataset_statistics}, and the details of the selected models are included in Appendix \ref{appendix models}.

\paragraph{Error Classification Taxonomies}
We consider four influential error classification taxonomies in error analysis: \textbf{POL73} \citep{linguistic_one}, \textbf{TUC74} \citep{gooficon}, \textbf{BRY17} \cite{errant}, and \textbf{FEI23} \citep{enhancing}. POL73 and TUC74 are linguistically driven but differ in hierarchical structure and classification logic. BRY17 introduces the rule-based ERRANT toolkit, categorizing errors by part-of-speech and token edit operations. FEI23, grounded in second language acquisition, adopts a cognitive perspective, classifying errors into single-word, inter-word, and discourse-level categories. Appendix \ref{appendix taxonomy} provides a detailed comparison of these taxonomies.

\begin{table}[tb]
\centering
\resizebox{0.42\textwidth}{!}{\begin{tabular}{l c c c c}
\toprule
\textbf{Annotator} & \textbf{POL73} & \textbf{TUC74} & \textbf{BRY17} & \textbf{FEI23} \\
\midrule
Annotator 1 \& 2 & 0.691 & 0.692 & \textbf{0.808} & 0.714 \\
Annotator 2 \& 3 & 0.813 & 0.661 & \textbf{0.851} & 0.824 \\
Annotator 1 \& 3 & 0.636 & 0.547 & \textbf{0.734} & 0.651 \\
\midrule
Average & 0.713 & 0.633 & \textbf{0.798} & 0.730 \\
\bottomrule
\end{tabular}}
\caption{Cohen’s Kappa Scores for Inter-Annotator Agreement across taxonomies.}
\label{table annotator}
\end{table}

\subsection{Main Results}

We systematically assess the rationality of various error classification taxonomies in Table \ref{tab1}.

\textbf{Exclusivity: Ambiguous type boundaries undermine mutual exclusivity.} Taxonomies with overlapping or poorly defined error categories lead to classification inconsistencies, making it difficult to assign a unique label to each error instance.

\textbf{Coverage: A greater number of categories does not guarantee comprehensive coverage.} Effective taxonomies must encompass both frequent and rare errors, as merely expanding categories does not guarantee comprehensive representation.

\textbf{Balance: Over-specification and Overgeneralization may disrupt distributional balance.} Over-specification introduces unnecessary fine-grained distinctions while overgeneralization clusters distinct errors into broad categories. Both lead to imbalanced error distributions.

\textbf{Usability: Linguistic-based taxonomies reduce practical utility.} Taxonomies based on intricate semantic or syntactic distinctions increase the complexity of computational modeling, limiting their usability in real-world applications.

Further analyses are provided in Appendix \ref{appendix main}.

\subsection{Annotator Agreement Analysis}
\label{section annotator agreement}
We measure inter-annotator agreement as a part of Usability, using Cohen’s Kappa Index \citep{cohen}, a standard metric for measuring annotation consistency. The results of three annotators are summarized in Table \ref{table annotator}.

BRY17 and FEI23 exhibit higher agreement scores compared to POL73 and TUC74, suggesting that the former taxonomies provide clearer and more well-defined categories. This aligns with the trend observed in Table \ref{tab1}, further substantiating the validity of our evaluation framework. Moreover, the higher agreement in BRY17 and FEI23 reinforces their practical applicability, while the lower scores in POL73 and TUC74 suggest potential ambiguity in category definitions.

To assess the impact of classification granularity, we conducted an ablation study by merging specific error categories and analyzing their effect on key metrics. The results underscore the need for rigorous validation of error classification taxonomies, rather than relying on intuition or convention, highlighting the importance of our proposed evaluation metrics. The detailed results are in Appendix \ref{appendix ablation study}.

\section{Conclusion}
We revisit error classification taxonomies in error analysis by proposing a systematic evaluation framework based on exclusivity, coverage, balance, and usability, evaluated on our own annotated dataset. Our experiments validate the effectiveness of these metrics and underscore the need for a systematic assessment of classification taxonomies. Results demonstrate that different taxonomies exhibit trade-offs, with some excelling in exclusivity and coverage while others offer better usability. These findings highlight the importance of well-structured taxonomies for reliable error analysis and annotation consistency.
\section*{Limitations}
One key limitation of this study is Exclusivity and Usability rely on large language models (LLMs) for computation. While our experimental results demonstrate the reasonableness of these metrics, LLMs inherently introduce biases that may lead to preferences for certain error classification frameworks over others. Future work could explore model-agnostic approaches to mitigate such biases.

Another limitation stems from the dataset scope. Our analysis is based on the W\&I+LOCNESS dataset, which, while widely used, may not fully capture the diversity of learner error patterns across different proficiency levels, native languages, and writing contexts. Extending our evaluation to more diverse datasets could improve the generalizability of our findings.
\section*{Ethics Statement}

We conduct our experiments using the publicly available W\&I+LOCNESS dataset, which does not contain sensitive data. All models used are also publicly available, and we have properly cited their sources. The datasets and models are utilized in accordance with their intended purposes.
For human agreement evaluation, we employed three postgraduate students specializing in foreign linguistics and applied linguistics as part-time annotators. Each annotator completed the entire annotation process within approximately 20 working hours and was compensated at a rate of \$50 per hour.

\bibliography{custom}

\begin{thebibliography}{54}
\providecommand{\natexlab}[1]{#1}

\bibitem[{AI@Meta(2024)}]{llama3modelcard}
AI@Meta. 2024.
\newblock \href {https://github.com/meta-llama/llama3/blob/main/MODEL_CARD.md} {Llama 3 model card}.

\bibitem[{Anthropic(2024)}]{claude3}
Anthropic. 2024.
\newblock \href {https://api.semanticscholar.org/CorpusID:268232499} {The claude 3 model family: Opus, sonnet, haiku}.

\bibitem[{Bertkua(1974)}]{other_one}
Jana~Svoboda Bertkua. 1974.
\newblock \href {https://doi.org/10.1111/j.1467-1770.1974.tb00508.x} {An analysis of english learner speech}.
\newblock \emph{Language Learning}, 24(2):279--286.

\bibitem[{Bialystok et~al.(1982)Bialystok, Dulay, Burt, and Krashen}]{language_two}
Ellen Bialystok, Heidi Dulay, Marina Burt, and Stephen Krashen. 1982.
\newblock \href {https://doi.org/10.2307/327086} {Language two}.
\newblock \emph{The Modern Language Journal}, 67:273.

\bibitem[{Bryant et~al.(2019)Bryant, Felice, Andersen, and Briscoe}]{w&i}
Christopher Bryant, Mariano Felice, {\O}istein~E. Andersen, and Ted Briscoe. 2019.
\newblock \href {https://doi.org/10.18653/v1/W19-4406} {The {BEA}-2019 shared task on grammatical error correction}.
\newblock In \emph{Proceedings of the Fourteenth Workshop on Innovative Use of NLP for Building Educational Applications}, pages 52--75, Florence, Italy. Association for Computational Linguistics.

\bibitem[{Bryant et~al.(2017)Bryant, Felice, and Briscoe}]{errant}
Christopher Bryant, Mariano Felice, and Ted Briscoe. 2017.
\newblock \href {https://doi.org/10.18653/v1/P17-1074} {Automatic annotation and evaluation of error types for grammatical error correction}.
\newblock In \emph{Proceedings of the 55th Annual Meeting of the Association for Computational Linguistics (Volume 1: Long Papers)}, pages 793--805, Vancouver, Canada. Association for Computational Linguistics.

\bibitem[{Burt(1975)}]{communicative_one}
Marina~K. Burt. 1975.
\newblock \href {http://www.jstor.org/stable/3586012} {Error analysis in the adult efl classroom}.
\newblock \emph{TESOL Quarterly}, 9(1):53--63.

\bibitem[{CHAN(2010)}]{other_two}
ALICE Y.~W. CHAN. 2010.
\newblock \href {https://doi.org/10.5054/tq.2010.219941} {Toward a taxonomy of written errors: Investigation into the written errors of hong kong cantonese esl learners}.
\newblock \emph{TESOL Quarterly}, 44(2):295--319.

\bibitem[{Cohen(1960)}]{cohen}
Jacob Cohen. 1960.
\newblock \href {https://doi.org/10.1177/001316446002000104} {A coefficient of agreement for nominal scales}.
\newblock \emph{Educational and Psychological Measurement}, 20(1):37--46.

\bibitem[{Corder(1967)}]{l13}
Stephen~Pit Corder. 1967.
\newblock The significance of learner's errors.

\bibitem[{Dahlmeier et~al.(2013)Dahlmeier, Ng, and Wu}]{NUCLE}
Daniel Dahlmeier, Hwee~Tou Ng, and Siew~Mei Wu. 2013.
\newblock \href {https://aclanthology.org/W13-1703/} {Building a large annotated corpus of learner {E}nglish: The {NUS} corpus of learner {E}nglish}.
\newblock In \emph{Proceedings of the Eighth Workshop on Innovative Use of {NLP} for Building Educational Applications}, pages 22--31, Atlanta, Georgia. Association for Computational Linguistics.

\bibitem[{Dodigovic(2007)}]{l11}
Marina Dodigovic. 2007.
\newblock \href {https://doi.org/10.2167/la416.0} {Artificial intelligence and second language learning: An efficient approach to error remediation}.
\newblock \emph{Language Awareness - LANG AWARE}, 16:99--113.

\bibitem[{Dulay and Burt(1972)}]{l10}
Heidi~C. Dulay and Marina~K. Burt. 1972.
\newblock \href {https://doi.org/10.1111/j.1467-1770.1972.tb00085.x} {Goofing: An indicator of children's second language learning strategies}.
\newblock \emph{Language Learning}, 22(2):235--252.

\bibitem[{Dulay and Burt(1974)}]{comparative_child}
Heidi~C. Dulay and Marina~K. Burt. 1974.
\newblock \href {http://www.jstor.org/stable/3585536} {Errors and strategies in child second language acquisition}.
\newblock \emph{TESOL Quarterly}, 8(2):129--136.

\bibitem[{Fei et~al.(2023)Fei, Cui, Yang, Lam, Lan, and Shi}]{enhancing}
Yuejiao Fei, Leyang Cui, Sen Yang, Wai Lam, Zhenzhong Lan, and Shuming Shi. 2023.
\newblock \href {https://doi.org/10.18653/v1/2023.acl-long.413} {Enhancing grammatical error correction systems with explanations}.
\newblock In \emph{Proceedings of the 61st Annual Meeting of the Association for Computational Linguistics (Volume 1: Long Papers)}, pages 7489--7501, Toronto, Canada. Association for Computational Linguistics.

\bibitem[{He et~al.(2021)He, Peng, Liao, Liu, and Xiong}]{TEGA}
Jie He, Bo~Peng, Yi~Liao, Qun Liu, and Deyi Xiong. 2021.
\newblock \href {https://doi.org/10.18653/v1/2021.acl-long.469} {{TGEA}: An error-annotated dataset and benchmark tasks for {T}ext{G}eneration from pretrained language models}.
\newblock In \emph{Proceedings of the 59th Annual Meeting of the Association for Computational Linguistics and the 11th International Joint Conference on Natural Language Processing (Volume 1: Long Papers)}, pages 6012--6025, Online. Association for Computational Linguistics.

\bibitem[{Heift and Schulze(2007)}]{l12}
Trude Heift and Mathias Schulze. 2007.
\newblock \href {https://doi.org/10.4324/9780203012215} {\emph{Errors and Intelligence in Computer-Assisted Language Learning: Parsers and Pedagogues}}, 1st edition.
\newblock Routledge.

\bibitem[{Huang et~al.(2023)Huang, Ye, Zhou, Li, Li, Zhou, and Zheng}]{huang-etal-2023-frustratingly}
Haojing Huang, Jingheng Ye, Qingyu Zhou, Yinghui Li, Yangning Li, Feng Zhou, and Hai-Tao Zheng. 2023.
\newblock \href {https://doi.org/10.18653/v1/2023.findings-emnlp.771} {A frustratingly easy plug-and-play detection-and-reasoning module for {C}hinese spelling check}.
\newblock In \emph{Findings of the Association for Computational Linguistics: EMNLP 2023}, pages 11514--11525, Singapore. Association for Computational Linguistics.

\bibitem[{James(1998)}]{l14}
Carl James. 1998.
\newblock \href {https://doi.org/10.4324/9781315842912} {\emph{Errors in Language Learning and Use: Exploring Error Analysis}}, 1st edition.
\newblock Routledge.

\bibitem[{Jiang et~al.(2023)Jiang, Sablayrolles, Mensch, Bamford, Chaplot, de~las Casas, Bressand, Lengyel, Lample, Saulnier, Lavaud, Lachaux, Stock, Scao, Lavril, Wang, Lacroix, and Sayed}]{mistral7b}
Albert~Q. Jiang, Alexandre Sablayrolles, Arthur Mensch, Chris Bamford, Devendra~Singh Chaplot, Diego de~las Casas, Florian Bressand, Gianna Lengyel, Guillaume Lample, Lucile Saulnier, Lélio~Renard Lavaud, Marie-Anne Lachaux, Pierre Stock, Teven~Le Scao, Thibaut Lavril, Thomas Wang, Timothée Lacroix, and William~El Sayed. 2023.
\newblock \href {https://arxiv.org/abs/2310.06825} {Mistral 7b}.
\newblock \emph{Preprint}, arXiv:2310.06825.

\bibitem[{Kafipour and Laleh(2012)}]{comparative_one}
Reza Kafipour and Khojasteh Laleh. 2012.
\newblock A comparative taxonomy of errors made by iranian undergraduate learners of english.
\newblock \emph{Canadian Social Science}, 8:18--24.

\bibitem[{Li et~al.(2023{\natexlab{a}})Li, Huang, Ma, Jiang, Li, Zhou, Zheng, and Zhou}]{li2023effectiveness}
Yinghui Li, Haojing Huang, Shirong Ma, Yong Jiang, Yangning Li, Feng Zhou, Hai-Tao Zheng, and Qingyu Zhou. 2023{\natexlab{a}}.
\newblock On the (in) effectiveness of large language models for chinese text correction.
\newblock \emph{arXiv preprint arXiv:2307.09007}.

\bibitem[{Li et~al.(2025)Li, Ma, Chen, Huang, Huang, Li, Zheng, and Shen}]{li2025correct}
Yinghui Li, Shirong Ma, Shaoshen Chen, Haojing Huang, Shulin Huang, Yangning Li, Hai-Tao Zheng, and Ying Shen. 2025.
\newblock Correct like humans: Progressive learning framework for chinese text error correction.
\newblock \emph{Expert Systems with Applications}, 265:126039.

\bibitem[{Li et~al.(2022{\natexlab{a}})Li, Ma, Zhou, Li, Yangning, Huang, Liu, Li, Cao, and Zheng}]{li2022learning}
Yinghui Li, Shirong Ma, Qingyu Zhou, Zhongli Li, Li~Yangning, Shulin Huang, Ruiyang Liu, Chao Li, Yunbo Cao, and Haitao Zheng. 2022{\natexlab{a}}.
\newblock Learning from the dictionary: Heterogeneous knowledge guided fine-tuning for chinese spell checking.
\newblock \emph{arXiv preprint arXiv:2210.10320}.

\bibitem[{Li et~al.(2024{\natexlab{a}})Li, Qin, Huang, Ye, Li, Qin, Hu, Jiang, Zheng, and Yu}]{li2024rethinkingroleslargelanguage}
Yinghui Li, Shang Qin, Haojing Huang, Jingheng Ye, Yangning Li, Libo Qin, Xuming Hu, Wenhao Jiang, Hai-Tao Zheng, and Philip~S. Yu. 2024{\natexlab{a}}.
\newblock \href {https://arxiv.org/abs/2402.11420} {Rethinking the roles of large language models in chinese grammatical error correction}.
\newblock \emph{Preprint}, arXiv:2402.11420.

\bibitem[{Li et~al.(2023{\natexlab{b}})Li, Xu, Chen, Huang, Li, Jiang, Li, Zhou, Zheng, and Shen}]{li2023towards}
Yinghui Li, Zishan Xu, Shaoshen Chen, Haojing Huang, Yangning Li, Yong Jiang, Zhongli Li, Qingyu Zhou, Hai-Tao Zheng, and Ying Shen. 2023{\natexlab{b}}.
\newblock Towards real-world writing assistance: A chinese character checking benchmark with faked and misspelled characters.
\newblock \emph{arXiv preprint arXiv:2311.11268}.

\bibitem[{Li et~al.(2022{\natexlab{b}})Li, Zhou, Li, Li, Liu, Sun, Wang, Li, Cao, and Zheng}]{li2022past}
Yinghui Li, Qingyu Zhou, Yangning Li, Zhongli Li, Ruiyang Liu, Rongyi Sun, Zizhen Wang, Chao Li, Yunbo Cao, and Hai-Tao Zheng. 2022{\natexlab{b}}.
\newblock The past mistake is the future wisdom: Error-driven contrastive probability optimization for chinese spell checking.
\newblock \emph{arXiv preprint arXiv:2203.00991}.

\bibitem[{Li et~al.(2024{\natexlab{b}})Li, Zhou, Luo, Ma, Li, Zheng, Hu, and Philip}]{li2024llms}
Yinghui Li, Qingyu Zhou, Yuanzhen Luo, Shirong Ma, Yangning Li, Hai-Tao Zheng, Xuming Hu, and S~Yu Philip. 2024{\natexlab{b}}.
\newblock When llms meet cunning texts: A fallacy understanding benchmark for large language models.
\newblock In \emph{The Thirty-eight Conference on Neural Information Processing Systems Datasets and Benchmarks Track}.

\bibitem[{Liang et~al.(2023)Liang, Davidson, Yuan, Panditharatne, Chen, Shea, Pham, Tan, Voss, and Fryer}]{l15}
Kai-Hui Liang, Sam Davidson, Xun Yuan, Shehan Panditharatne, Chun-Yen Chen, Ryan Shea, Derek Pham, Yinghua Tan, Erik Voss, and Luke Fryer. 2023.
\newblock \href {https://doi.org/10.18653/v1/2023.bea-1.7} {{C}hat{B}ack: Investigating methods of providing grammatical error feedback in a {GUI}-based language learning chatbot}.
\newblock In \emph{Proceedings of the 18th Workshop on Innovative Use of NLP for Building Educational Applications (BEA 2023)}, pages 83--99, Toronto, Canada. Association for Computational Linguistics.

\bibitem[{Liu et~al.(2022)Liu, Li, Tao, Liang, and Zheng}]{liu2022we}
Ruiyang Liu, Yinghui Li, Linmi Tao, Dun Liang, and Hai-Tao Zheng. 2022.
\newblock Are we ready for a new paradigm shift? a survey on visual deep mlp.
\newblock \emph{Patterns}, 3(7).

\bibitem[{Ma et~al.(2022)Ma, Li, Sun, Zhou, Huang, Zhang, Yangning, Liu, Li, Cao et~al.}]{ma2022linguistic}
Shirong Ma, Yinghui Li, Rongyi Sun, Qingyu Zhou, Shulin Huang, Ding Zhang, Li~Yangning, Ruiyang Liu, Zhongli Li, Yunbo Cao, et~al. 2022.
\newblock Linguistic rules-based corpus generation for native chinese grammatical error correction.
\newblock \emph{arXiv preprint arXiv:2210.10442}.

\bibitem[{Nicholls(2003)}]{other_four}
Diane Nicholls. 2003.
\newblock The cambridge learner corpus: Error coding and analysis for lexicography and elt.
\newblock In \emph{Proceedings of the Corpus Linguistics 2003 conference}, volume~16, pages 572--581. Cambridge University Press Cambridge.

\bibitem[{OpenAI(2024)}]{chatgpt4o}
OpenAI. 2024.
\newblock \href {https://arxiv.org/abs/2410.21276} {Gpt-4o system card}.
\newblock \emph{Preprint}, arXiv:2410.21276.

\bibitem[{{\"O}zkayran and Yılmaz(2020)}]{surface_one}
Ali {\"O}zkayran and Emrullah Yılmaz. 2020.
\newblock \href {https://api.semanticscholar.org/CorpusID:221868114} {Analysis of higher education students’ errors in english writing tasks}.
\newblock \emph{Advances in Language and Literary Studies}.

\bibitem[{Politzer and Ramirez(1973)}]{linguistic_one}
Robert~L. Politzer and Arnulfo~G. Ramirez. 1973.
\newblock \href {https://doi.org/10.1111/j.1467-1770.1973.tb00096.x} {An error analysis of the spoken english of mexican-american pupils in a bilingual school and a monolingual school}.
\newblock \emph{Language Learning}, 23(1):39--61.

\bibitem[{Rixha et~al.(2021)Rixha, Alhamid, Rokhmah, and Ukka}]{surface_two}
Annafi’in~Nur Rixha, Idrus Alhamid, Siti Rokhmah, and Syamsir~Bin Ukka. 2021.
\newblock \href {https://doi.org/10.53491/kariwarismart.v1i2.39} {Surface strategy taxonomy: Grammatical errors analysis in the third-semester students’ descriptive essay}.
\newblock \emph{KARIWARI SMART : Journal of Education Based on Local Wisdom}, 1(2):36--48.

\bibitem[{Suhono(2017)}]{surface_three}
Suhono Suhono. 2017.
\newblock \href {https://doi.org/10.25217/ji.v1i2.128} {Surface strategy taxonomy on the efl students composition a study of error analysis}.
\newblock 1:1–30.

\bibitem[{Tucker et~al.(1974)Tucker, Burt, and Kiparsky}]{gooficon}
G~R Tucker, M~K Burt, and C~Kiparsky. 1974.
\newblock \href {https://doi.org/10.2307/3585545} {The gooficon: A repair manual for english}.
\newblock \emph{TESOL Quarterly}, 8(2):191.

\bibitem[{Törnberg(2024)}]{annotation_one}
Petter Törnberg. 2024.
\newblock \href {https://doi.org/10.6092/issn.1971-8853/19461} {Best practices for text annotation with large language models}.
\newblock \emph{Sociologica}, 18(2):67–85.

\bibitem[{Waruwu and Harefa(2024)}]{communicative_two}
Yaredi Waruwu and Afore~Tahir Harefa. 2024.
\newblock \href {https://doi.org/10.55637/jr.10.2.10226.626-636} {An error analysis of communicative effect taxonomy of the tenth grade students’ writing of descriptive text}.
\newblock \emph{RETORIKA: Jurnal Ilmu Bahasa}, 10(2):626–636.

\bibitem[{White and Ontario Institute for Studies~in Education(1977)}]{comparative_adult}
Lydia White and Toronto. Bilingual Education~Project Ontario Institute for Studies~in Education. 1977.
\newblock \href {https://nla.gov.au/nla.cat-vn5337347} {Error analysis and error correction in adult learners of english as a second language. working papers on bilingualism, no. 13 [microform]}.
\newblock Accessed: 09 February 2025.

\bibitem[{Xiong et~al.(2024)Xiong, Hu, Lu, Li, Fu, He, and Hooi}]{confidence}
Miao Xiong, Zhiyuan Hu, Xinyang Lu, Yifei Li, Jie Fu, Junxian He, and Bryan Hooi. 2024.
\newblock \href {https://arxiv.org/abs/2306.13063} {Can llms express their uncertainty? an empirical evaluation of confidence elicitation in llms}.
\newblock \emph{Preprint}, arXiv:2306.13063.

\bibitem[{Yannakoudakis et~al.(2017)Yannakoudakis, Rei, Andersen, and Yuan}]{l16}
Helen Yannakoudakis, Marek Rei, {\O}istein~E. Andersen, and Zheng Yuan. 2017.
\newblock \href {https://doi.org/10.18653/v1/D17-1297} {Neural sequence-labelling models for grammatical error correction}.
\newblock In \emph{Proceedings of the 2017 Conference on Empirical Methods in Natural Language Processing}, pages 2795--2806, Copenhagen, Denmark. Association for Computational Linguistics.

\bibitem[{Ye et~al.(2023{\natexlab{a}})Ye, Li, Li, and Zheng}]{ye-etal-2023-mixedit}
Jingheng Ye, Yinghui Li, Yangning Li, and Hai-Tao Zheng. 2023{\natexlab{a}}.
\newblock \href {https://doi.org/10.18653/v1/2023.findings-emnlp.681} {{M}ix{E}dit: Revisiting data augmentation and beyond for grammatical error correction}.
\newblock In \emph{Findings of the Association for Computational Linguistics: EMNLP 2023}, pages 10161--10175, Singapore. Association for Computational Linguistics.

\bibitem[{Ye et~al.(2022)Ye, Li, Ma, Xie, Wu, and Zheng}]{ye2022focus}
Jingheng Ye, Yinghui Li, Shirong Ma, Rui Xie, Wei Wu, and Hai-Tao Zheng. 2022.
\newblock Focus is what you need for chinese grammatical error correction.
\newblock \emph{arXiv preprint arXiv:2210.12692}.

\bibitem[{Ye et~al.(2023{\natexlab{b}})Ye, Li, and Zheng}]{ye-etal-2023-system}
Jingheng Ye, Yinghui Li, and Haitao Zheng. 2023{\natexlab{b}}.
\newblock \href {https://aclanthology.org/2023.ccl-3.29/} {System report for {CCL}23-eval task 7: {THU} {KEL}ab (sz) - exploring data augmentation and denoising for {C}hinese grammatical error correction}.
\newblock In \emph{Proceedings of the 22nd Chinese National Conference on Computational Linguistics (Volume 3: Evaluations)}, pages 262--270, Harbin, China. Chinese Information Processing Society of China.

\bibitem[{Ye et~al.(2023{\natexlab{c}})Ye, Li, Zhou, Li, Ma, Zheng, and Shen}]{ye-etal-2023-cleme}
Jingheng Ye, Yinghui Li, Qingyu Zhou, Yangning Li, Shirong Ma, Hai-Tao Zheng, and Ying Shen. 2023{\natexlab{c}}.
\newblock \href {https://doi.org/10.18653/v1/2023.emnlp-main.378} {{CLEME}: Debiasing multi-reference evaluation for grammatical error correction}.
\newblock In \emph{Proceedings of the 2023 Conference on Empirical Methods in Natural Language Processing}, pages 6174--6189, Singapore. Association for Computational Linguistics.

\bibitem[{Ye et~al.(2024{\natexlab{a}})Ye, Qin, Li, Cheng, Qin, Zheng, Xing, Xu, Cheng, and Wei}]{excgec}
Jingheng Ye, Shang Qin, Yinghui Li, Xuxin Cheng, Libo Qin, Hai-Tao Zheng, Peng Xing, Zishan Xu, Guo Cheng, and Zhao Wei. 2024{\natexlab{a}}.
\newblock \href {https://arxiv.org/abs/2407.00924} {Excgec: A benchmark of edit-wise explainable chinese grammatical error correction}.
\newblock \emph{Preprint}, arXiv:2407.00924.

\bibitem[{Ye et~al.(2025)Ye, Wang, Zou, Yan, Wang, Zheng, Xu, King, Yu, and Wen}]{ye2025position}
Jingheng Ye, Shen Wang, Deqing Zou, Yibo Yan, Kun Wang, Hai-Tao Zheng, Zenglin Xu, Irwin King, Philip~S Yu, and Qingsong Wen. 2025.
\newblock Position: Llms can be good tutors in foreign language education.
\newblock \emph{arXiv preprint arXiv:2502.05467}.

\bibitem[{Ye et~al.(2024{\natexlab{b}})Ye, Xu, Li, Cheng, Song, Zhou, Zheng, Shen, and Su}]{ye2024cleme2}
Jingheng Ye, Zishan Xu, Yinghui Li, Xuxin Cheng, Linlin Song, Qingyu Zhou, Hai-Tao Zheng, Ying Shen, and Xin Su. 2024{\natexlab{b}}.
\newblock Cleme2. 0: Towards more interpretable evaluation by disentangling edits for grammatical error correction.
\newblock \emph{arXiv preprint arXiv:2407.00934}.

\bibitem[{Zhang et~al.(2024)Zhang, Almpanidis, Fan, Deng, Zhang, Liu, Kamel, Soda, and Gama}]{lt2}
Chongsheng Zhang, George Almpanidis, Gaojuan Fan, Binquan Deng, Yanbo Zhang, Ji~Liu, Aouaidjia Kamel, Paolo Soda, and João Gama. 2024.
\newblock \href {https://arxiv.org/abs/2408.00483} {A systematic review on long-tailed learning}.
\newblock \emph{Preprint}, arXiv:2408.00483.

\bibitem[{Zhang et~al.(2023{\natexlab{a}})Zhang, Li, Zhou, Ma, Li, Cao, and Zheng}]{zhang2023contextual}
Ding Zhang, Yinghui Li, Qingyu Zhou, Shirong Ma, Yangning Li, Yunbo Cao, and Hai-Tao Zheng. 2023{\natexlab{a}}.
\newblock Contextual similarity is more valuable than character similarity: An empirical study for chinese spell checking.
\newblock In \emph{ICASSP 2023-2023 IEEE International Conference on Acoustics, Speech and Signal Processing (ICASSP)}, pages 1--5. IEEE.

\bibitem[{Zhang et~al.(2023{\natexlab{b}})Zhang, Kang, Hooi, Yan, and Feng}]{lt1}
Yifan Zhang, Bingyi Kang, Bryan Hooi, Shuicheng Yan, and Jiashi Feng. 2023{\natexlab{b}}.
\newblock \href {https://arxiv.org/abs/2110.04596} {Deep long-tailed learning: A survey}.
\newblock \emph{Preprint}, arXiv:2110.04596.

\bibitem[{Ângela Costa et~al.(2015)Ângela Costa, Ling, Luís, Correia, and Coheur}]{linguistic_three}
Ângela Costa, Wang Ling, Tiago Luís, Rui Correia, and Luísa Coheur. 2015.
\newblock \href {http://www.jstor.org/stable/44113801} {A linguistically motivated taxonomy for machine translation error analysis}.
\newblock \emph{Machine Translation}, 29(2):127--161.

\end{thebibliography}

\appendix

\section{Notation Table}
\label{appendix notation table}
\begin{table*}[htb]
    \centering
    \resizebox{0.9\textwidth}{!}{
    \begin{tabular}{c c}
        \toprule
        \textbf{Symbol} & \textbf{Description} \\ 
        \midrule
        $C^{(x)}$  & The placement confidence of sample $x$ \\ 
        $C^{(x)}_i$ & Placement confidence of the $i$-th predicted error type of sample $x$ in the top-k setting \\ 
        $\hat{Y}^{(x)}$ & Predicted error type of sample $x$ \\ 
        $\hat{Y}^{(x)}_i$ & The $i$-th predicted error type of sample $x$ in the top-k setting \\ 
        $\tilde{Y}$ & Gold reference of sample $x$ \\ 
        $F$ & Error classification taxonomy \\ 
        $F_i$ & The $i$-th error classification taxonomy \\ 
        $n$ & Number of error classification taxonomies \\ 
        $ET_j$ & The $j$-th error type \\ 
        $m$ & Number of error types in the corresponding error classification taxonomy $F$ \\
        $\tau$ & the predefined confidence threshold \\
        $k$ & the selection parameter in the Top-K Prompting Strategy \\
        \bottomrule
    \end{tabular}}
    \caption{Notation Table.}
    \label{notation table}
\end{table*}

As shown in Table \ref{notation table}, this appendix provides a list of relevant symbols and their meanings for reference.

\section{Strategies for Robust Confidence Estimation}
\label{appendix confidence}
To enhance the robustness of \textit{confidence estimation}, we employ three strategies \citep{confidence} that mitigate overconfidence and improve the reliability of confidence scores:

\paragraph{Top-K Prompting Strategy.} One effective approach to reducing overconfidence in LLMs is recognizing the existence of multiple plausible answers. To incorporate this principle, we modify the prompt to explicitly request the generation of the top 3 distinct error types along with their respective confidence scores. By enforcing diversity among the generated error types, this strategy encourages a more calibrated confidence estimation by considering multiple reasonable interpretations.

\paragraph{Self-random Sampling Strategy.} Analyzing the variation among multiple responses to the same input can further aid in confidence calibration. This strategy leverages the inherent randomness of LLMs by issuing the same prompt multiple times, thereby introducing controlled variability in the responses. To ensure sufficient diversity in the generated results, we set the temperature to 0.7 across all model configurations.

\paragraph{Average-Confidence (Avg-Conf) Aggregation Strategy.} Once multiple responses are generated, an aggregation mechanism is necessary to effectively capture and utilize the observed variability. We adopt the Avg-Conf aggregation strategy, which computes the final confidence distribution by averaging the confidence scores across multiple samples. This approach enables a more robust estimation of confidence levels by accounting for response fluctuations.

By integrating these strategies, we enable LLMs to articulate confidence scores more precisely, leading to improved robustness and trustworthiness in decision-making. For more details, please refer to \citet{confidence}.

\section{Dataset and Annotation Details}
\subsection{Data Preprocessing}
\label{appendix extract single}
The original dataset contains numerous sentences with $n \ (n>1)$ grammatical errors. If such sentences are directly input into the model for error classification, it introduces ambiguity—making it unclear which error the model should focus on. This uncertainty can negatively impact prediction consistency. To mitigate this issue, we preprocess the dataset by isolating individual grammatical errors. Specifically, for each sentence containing $n$ errors, we generate $n$ separate instances using the edit extraction toolkit ERRANT~\cite{errant} and CLEME \cite{ye-etal-2023-cleme,ye2024cleme2}, each retaining only one grammatical error while correcting the others. This ensures that each model input is unambiguous, leading to more reliable classification results.

\begin{table}[tb]
    \centering
    \resizebox{0.42\textwidth}{!}{
    \begin{tabular}{lccc}
        \toprule
        Attribute & Value \\
        \midrule
        Total Samples & 487 \\
        Edits per Sample & 1 \\
        Annotation taxonomies & POL73, TUC74, BRY17, FEI23 \\
        \bottomrule
    \end{tabular}}
    \caption{Statistics of the dataset.}
    \label{tab dataset_statistics}
\end{table}

\subsection{Annotation Approach}
\label{appendix annotation}
In this process, we strictly adhered to the standards and best practices outlined in \citet{annotation_one} to ensure that LLM-based annotations are reliable, reproducible, and ethical. First, we utilize the LLM for pre-annotation to reduce the manual workload and improve annotation efficiency. LLM-generated annotations are then reviewed by the authors through an annotation refinement process, where prompts are iteratively refined and optimized to ensure that the LLM fully understands the task requirements.  Following the first review, two professional linguists conducted a second round of evaluation. Given the first-round labels, they re-evaluated the data. If discrepancies arose between the linguists and the initial annotation, they engaged in discussions with the authors to finalize the label. This multi-step annotation workflow ensured the reliability and consistency of our dataset.

\subsection{Dataset Details}
\label{appendix dataset_statistics}
Our dataset consists of 487 samples, each containing a single edit. Each sample is annotated with four labels corresponding to different error classification taxonomies: POL73, TUC74, BRY17, and FEI23. The original dataset does not contain any sensitive personal data, and all data has been anonymized to ensure privacy. Table \ref{tab dataset_statistics} provides an overview of the dataset statistics.

\section{Model Details}
\label{appendix models}
In this study, we employed multiple large language models (LLMs) to evaluate the rationality of different error classification taxonomies in error analysis: \textbf{Meta-Llama-3-8B-Instruct} \citep{llama3modelcard}, \textbf{Mistral-7B-Instruct-v0.2} \citep{mistral7b}, \textbf{Claude-3-Haiku} \citep{claude3}, and \textbf{ChatGPT-4o} \citep{chatgpt4o}. These models were selected for their advanced text comprehension and classification capabilities, particularly in handling complex linguistic structures. Meta-Llama-3-8B-Instruct and Mistral-7B-Instruct-v0.2 were deployed on a local server equipped with an A800 GPU for inference, while Claude-3-Haiku and ChatGPT-4o were accessed via API calls.

\section{Error Classification Taxonomy Details}
\label{appendix taxonomy}
\subsection{POL73}
\label{subsec: linguistic one}
\begin{figure*}[htbp]
    \centering
    \includegraphics[width=\textwidth]{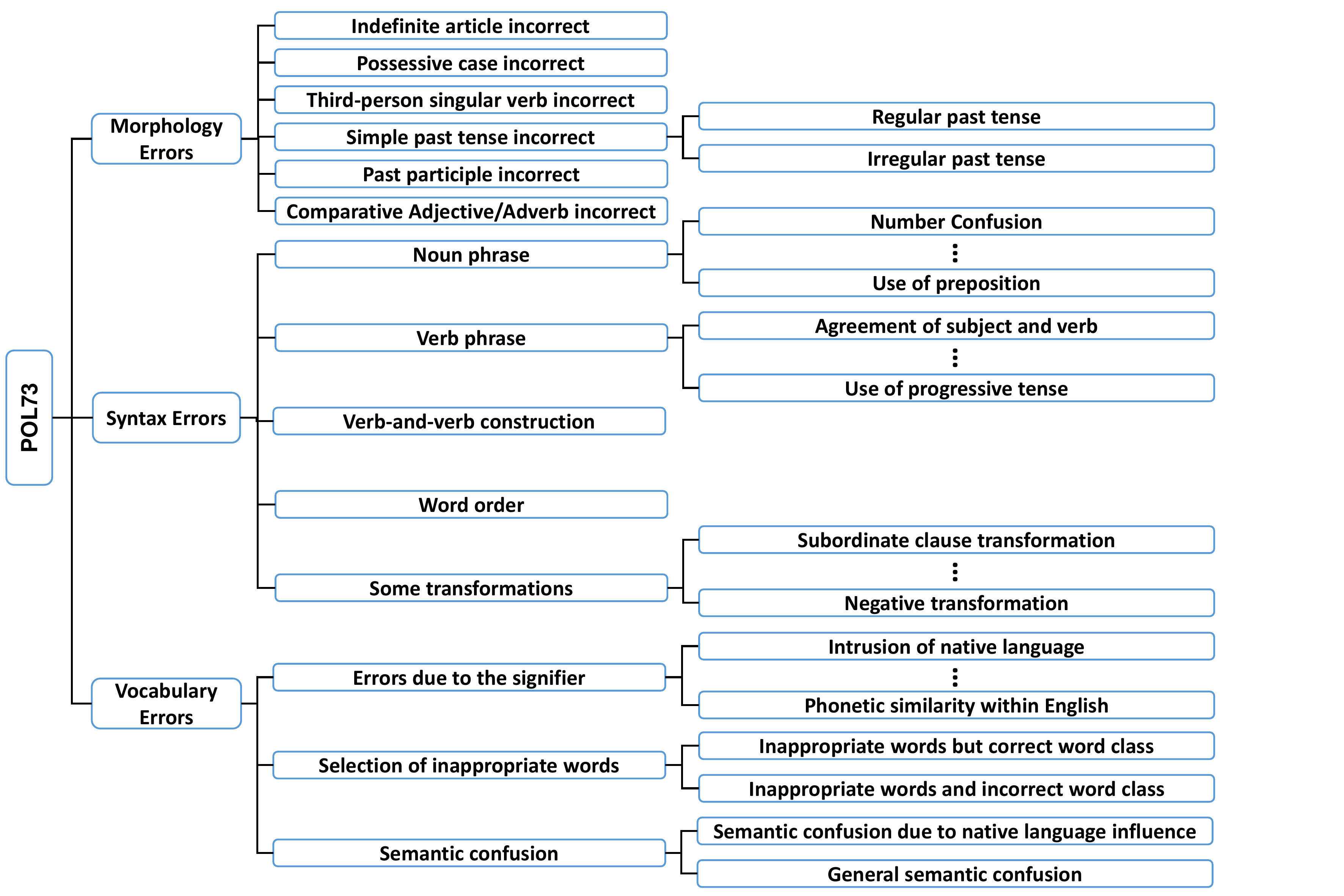}
    \caption{Overview of the POL73 Error Classification Taxonomy. The vertical ellipsis indicates that the category has additional subcategories not fully expanded here.}
    \label{fig:pol}
\end{figure*}

This hierarchical error taxonomy, based on a traditional descriptive taxonomy in linguistics, divides errors into three main categories: \textbf{Morphology}, \textbf{Syntax}, and \textbf{Vocabulary}.
Each category contains subcategories with definitions and examples for clarity. Figure \ref{fig:pol} provides an overview of this taxonomy. Below, we provide a detailed description of this taxonomy:

\begin{tcolorbox}[breakable]
\noindent \textbf{1 Morphology Errors}\\
\textbf{Definition:} Errors related to the form or structure of words, often involving affixes, tense, agreement, or comparative forms.

\noindent \textbf{1.1 Indefinite Article Incorrect}\\
\textbf{Definition:} Errors in the use of "a" or "an", such as using "a" before vowels or "an" before consonants.

\textbf{Source}: He saw an apple tree with \textcolor{red}{a} egg. \\
\textbf{Target}: He saw an apple tree with \textcolor{red}{an} egg.

\noindent \textbf{1.2 Possessive Case Incorrect}\\
\textbf{Definition:} Errors in forming possessives, including omission, misuse, or incorrect placement of apostrophes.

\textbf{Source}: \textcolor{red}{Mother’s Linda} came to visit. \\
\textbf{Target}: \textcolor{red}{Linda’s mother} came to visit.

\noindent \textbf{1.3 Third-person Singular Verb Incorrect}\\
\textbf{Definition:} Errors in using third-person singular verb forms, including omission or incorrect addition of "s" or "es."

\textbf{Source}: She \textcolor{red}{walk} to school every day. \\
\textbf{Target}: She \textcolor{red}{walks} to school every day.

\noindent \textbf{1.4 Simple Past Tense Incorrect}\\
\textbf{Definition:} Errors in forming the simple past tense for both regular and irregular verbs.

\noindent  \textbf{1.4.1 Regular Past Tense}\\
\textbf{Definition:} Errors involving omission or incorrect addition of "-ed" to regular verbs.

\textbf{Source}: He \textcolor{red}{walk} to the store. \\
\textbf{Target}: He \textcolor{red}{walked} to the store.

\noindent  \textbf{1.4.2 Irregular Past Tense}\\
\textbf{Definition:} Errors involving incorrect formation of the past tense for irregular verbs.

\textbf{Source}: He \textcolor{red}{goed} to the park. \\
\textbf{Target}: He \textcolor{red}{went} to the park.

\noindent \textbf{1.5 Past participle incorrect}\\
\textbf{Definition:} Errors in forming or using past participles correctly.

\textbf{Source}: He was \textcolor{red}{call}. \\
\textbf{Target}: He was \textcolor{red}{called}.

\noindent \textbf{1.6 Comparative Adjective/Adverb incorrect}\\
\textbf{Definition:} Errors in forming or using comparative forms of adjectives and adverbs.

\textbf{Source}: This house is \textcolor{red}{more} bigger. \\
\textbf{Target}: This house is bigger.

\noindent \textbf{2 Syntax Errors}\\
\textbf{Definition:} Errors related to sentence structure, including issues with phrases, word order, and transformations.

\noindent \textbf{2.1 Noun Phrase}\\
\textbf{Definition:} Errors in constructing or modifying a noun phrase.

\noindent \textbf{2.1.1 Determiners}\\
\textbf{Definition:} Errors in the omission, selection, or placement of determiners.

\textbf{Source}: He bought \textcolor{red}{the} apple from a market. \\
\textbf{Target}: He bought \textcolor{red}{an} apple from a market.

\noindent  \textbf{2.1.2 Nominalization}\\
\textbf{Definition:} Errors in using nominalized forms, such as substituting a base verb or incorrect structure for the "-ing" form in appropriate contexts.

\textbf{Source}: He improved the dish by \textcolor{red}{to cook} it longer. \\
\textbf{Target}: He improved the dish by \textcolor{red}{cooking} it longer.

\noindent \textbf{2.1.3 Number Confusion}\\
\textbf{Definition:} Misuse of singular or plural forms.

\textbf{Source}: She has many \textcolor{red}{friend}. \\
\textbf{Target}: She has many \textcolor{red}{friends}.

\noindent \textbf{2.1.4 Use of Pronoun}\\
\textbf{Definition:} Errors in pronoun reference, form, or usage, including omission, redundancy, substitution, or incorrect agreement with antecedents.

\textbf{Source}: \textcolor{red}{Her} went to the store. \\
\textbf{Target}: \textcolor{red}{She} went to the store.

\noindent \textbf{2.1.5 Use of Preposition}\\
\textbf{Definition:} Errors involving the misuse, omission, or substitution of prepositions in relation to verbs, noun phrases, or expressions of location and direction.

\textbf{Source}: He is good \textcolor{red}{in} math. \\
\textbf{Target}: He is good \textcolor{red}{at} math.

\noindent \textbf{2.2 Verb Phrase}\\
\textbf{Definition:} Errors related to verb use, including tense, form, and omission.

\noindent \textbf{2.2.1 Omission of Verb}\\
\textbf{Definition:} Errors involving the omission of a necessary verb, including main verbs and auxiliary verbs such as forms of "be".

\textbf{Source}: He to school yesterday. \\
\textbf{Target}: He \textcolor{red}{went} to school yesterday.

\noindent \textbf{2.2.2 Use of Progressive Tense}\\
\textbf{Definition:} Errors in forming or using the progressive tense, including omission of the auxiliary verb "be", incorrect use of the "-ing" form, or substitution of the progressive where it is not expected.

\textbf{Source}: He is \textcolor{red}{play} football. \\
\textbf{Target}: He is \textcolor{red}{playing} football.

\noindent \textbf{2.2.3 Agreement of Subject and Verb}\\
\textbf{Definition:} Errors where the subject and verb fail to agree in number, person, or tense.

\textbf{Source}: I didn't know what it \textcolor{red}{is}. \\
\textbf{Target}: I didn't know what it \textcolor{red}{was}.

\noindent \textbf{2.3 Verb-and-Verb Construction}\\
\textbf{Definition:} Errors in embedding one verb phrase into another, often involving omissions or misuse of linking elements like "to" or incorrect verb forms.

\textbf{Source}: She wants \textcolor{red}{goes} home. \\
\textbf{Target}: She wants \textcolor{red}{to go} home.

\noindent \textbf{2.4 Word Order}\\
\textbf{Definition:} Errors in the logical arrangement of words in a sentence, including incorrect placement of subjects, verbs, objects, or modifiers.

\textbf{Source}: \textcolor{red}{To school she goes} every day. \\
\textbf{Target}: \textcolor{red}{She goes to school} every day.

\noindent \textbf{2.5 Some Transformations}\\
\textbf{Definition:} Errors in applying specific grammatical transformations, such as passive voice, negations, questions, or subordinate clauses.

\noindent \textbf{2.5.1 Passive Transformation}\\
\textbf{Definition:} Errors in the formation or structure of passive voice sentences.

\textbf{Source}: The bird \textcolor{red}{got} saved. \\
\textbf{Target}: The bird \textcolor{red}{was} saved.

\noindent \textbf{2.5.2 Negative Transformation}\\
\textbf{Definition:} Errors in constructing negative sentences, including misuse of auxiliary verbs or double negations.

\textbf{Source}: She doesn't \textcolor{red}{likes} the movie. \\
\textbf{Target}: She doesn't \textcolor{red}{like} the movie.

\noindent \textbf{2.5.3 Question Transformation}\\
\textbf{Definition:} Errors in forming questions, often involving incorrect word order or auxiliary usage.

\textbf{Source}: Where \textcolor{red}{he is} going? \\
\textbf{Target}: Where \textcolor{red}{is he} going?

\noindent \textbf{2.5.4 There Transformation}\\
\textbf{Definition:} Errors in using "there" as a subject, including misuse or omission in existential constructions.

\textbf{Source}: There \textcolor{red}{is} many people in the room. \\
\textbf{Target}: There \textcolor{red}{are} many people in the room.

\noindent \textbf{2.5.5 Subordinate Clause Transformation}\\
\textbf{Definition:} Errors in constructing subordinate clauses, such as unnecessary elements or incorrect conjunction usage.

\textbf{Source}: I know \textcolor{red}{that} where she lives. \\
\textbf{Target}: I know where she lives.

\noindent \textbf{3 Vocabulary Errors}\\
\textbf{Definition:} Errors caused by confusion in word meaning, selection, or form, often due to similar sounds, inappropriate use in grammatical constructions, or meaning substitution.

\noindent \textbf{3.1 Errors Due to the Signifier}\\
\textbf{Definition:} The form or sound of a word leads to a misinterpretation or incorrect usage in a new linguistic context.

\noindent \textbf{3.1.1 Intrusion of Native Language}\\
\textbf{Definition:} Errors resulting from the direct influence of the speaker's native language on English usage.

\textbf{Source}: He \textcolor{red}{no lo mat}. \\
\textbf{Target}: He \textcolor{red}{does not kill it}

\noindent \textbf{3.1.2 Phonetic Similarities Between Native Language and English}\\
\textbf{Definition:} Errors caused by words resembling native language phonetics but incorrect in English.

\textbf{Source}: The \textcolor{red}{parablem} was difficult to solve. \\
\textbf{Target}: The \textcolor{red}{problem} was difficult to solve.

\noindent \textbf{3.1.3 Phonetic Similarity within English}\\
\textbf{Definition:} Errors stemming from mishearing or phonetic misinterpretation of English words.

\textbf{Source}: He tried to \textcolor{red}{kid} the mosquito. \\
\textbf{Target}: He tried to \textcolor{red}{kill} the mosquito.

\noindent \textbf{3.1.4 New Creations}\\
\textbf{Definition:} Errors caused by inventing non-existent English words for unclear reasons.

\textbf{Source}: He was \textcolor{red}{drownding} in the water. \\
\textbf{Target}: He was \textcolor{red}{drowning} in the water.

\noindent \textbf{3.2 Selection of Inappropriate Words}\\
\textbf{Definition:} Errors where a word with an incorrect meaning or grammatical role is chosen, despite the sentence being grammatically constructed correctly.

\noindent  \textbf{3.2.1 Inappropriate Words But Correct Word Class}\\
\textbf{Definition:} Errors where the word belongs to the correct grammatical class but is semantically inappropriate.

\textbf{Source}: They won't help \textcolor{red}{together}. \\
\textbf{Target}: They won't help \textcolor{red}{each other}.

\noindent  \textbf{3.2.2 Inappropriate Words and Incorrect Word Class}\\
\textbf{Definition:} Errors where the word is both grammatically and semantically inappropriate.

\textbf{Source}: The ant has \textcolor{red}{an open} in the ground. \\
\textbf{Target}: The ant has \textcolor{red}{a hole} in the ground.

\noindent \textbf{3.3 Semantic Confusion}\\
\textbf{Definition:} Errors occur when a word or phrase is used incorrectly due to misunderstanding its meaning or the influence of another language.

\noindent \textbf{3.3.1 Semantic Confusion Due to Native Language Influence}\\
\textbf{Definition:} Errors caused by incorrect meaning or usage influenced by the native language.

\textbf{Source}: The little ant \textcolor{red}{was} to the house. [note: (Spanish, fue) == (English, was) == (English, were)] \\
\textbf{Target}: The little ant \textcolor{red}{went} to the house.

\noindent \textbf{3.3.2 General Semantic Confusion}\\
\textbf{Definition:} Errors caused by inappropriate word usage that cannot be traced to native language influence.

\textbf{Source}: The man \textcolor{red}{came} into the water. \\
\textbf{Target}: The man \textcolor{red}{fell} into the water.

\end{tcolorbox}

\subsection{TUC74}
\label{subsec: gooficon}
\begin{figure*}[htbp]
    \centering
    \includegraphics[width=\textwidth]{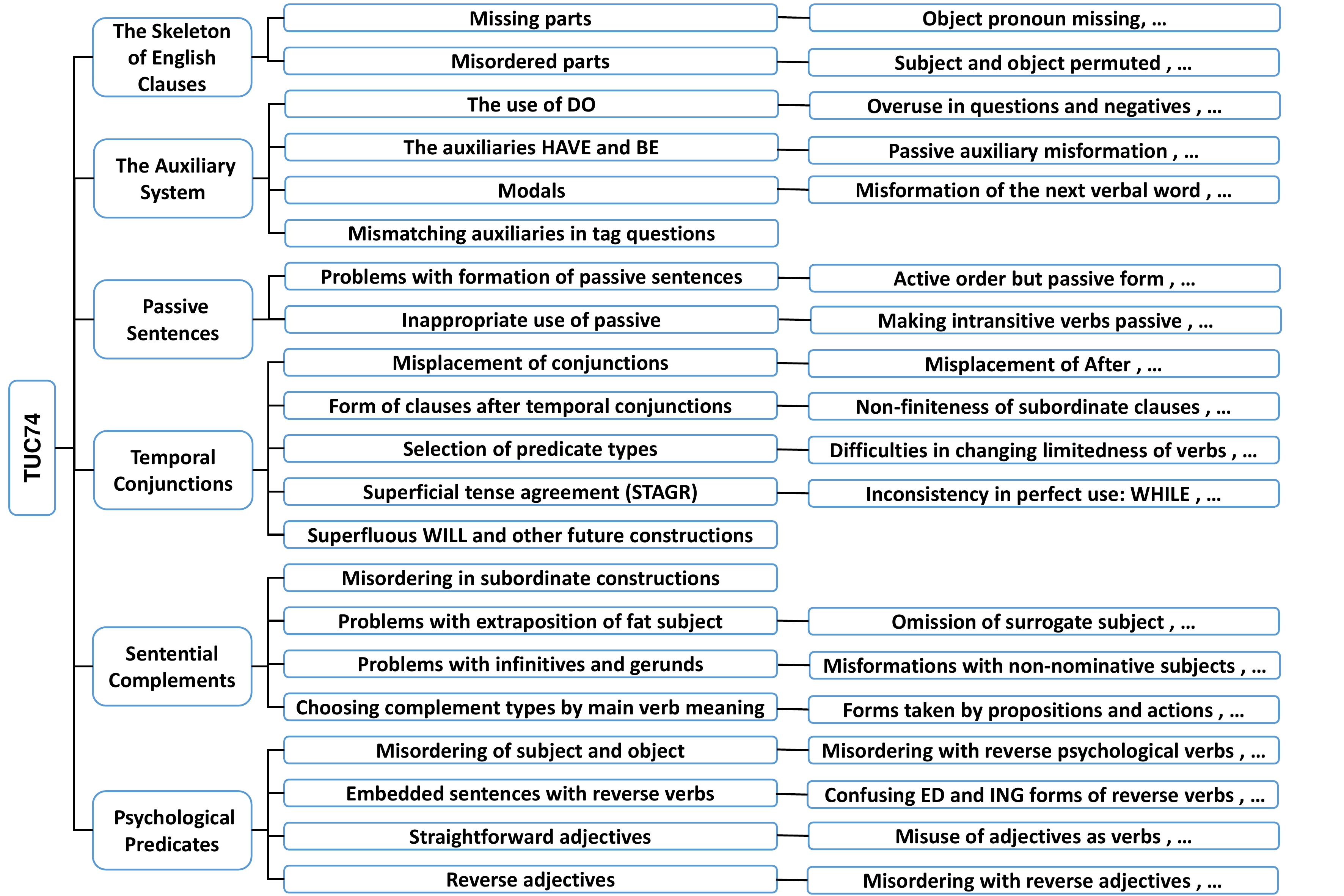}
    \caption{Overview of the TUC74 Error Classification Taxonomy. The horizontal ellipsis indicates that the category has additional subcategories not fully expanded here.}
    \label{fig:tuc}
\end{figure*}

This error taxonomy is a hierarchical taxonomy based on the structural, functional, and grammatical characteristics of English, encompassing errors in syntax, morphology, and semantic alignment within sentence construction.
Each category contains subcategories with definitions and examples for clarity. Figure \ref{fig:tuc} provides an overview of this taxonomy. Below, we provide a detailed description of this taxonomy:

\begin{tcolorbox}[breakable]
\noindent \textbf{1 The Skeleton of English Clauses}\\
\textbf{Definition:} Fundamental grammatical errors that compromise the basic structure of English clauses, including omissions and misplacements of critical components.

\noindent \textbf{1.1 Missing Parts}\\
\textbf{Definition:} Errors involving the omission of essential sentence components necessary for grammatical completeness and structural clarity.

\noindent \textbf{1.1.1 Surrogate Subject Missing: there and it}\\
\textbf{Definition:} Missing placeholder subjects "there" or "it," which are required in English sentence structure.

\textbf{Source}: \textcolor{red}{Was} a riot last night. \\
\textbf{Target}: \textcolor{red}{There was} a riot last night.

\noindent \textbf{1.1.2 Simple Predicate Missing: \textit{be}}\\
\textbf{Definition:} Missing the simple predicate "be," which is required when the predicate consists of adjectives or noun phrases.

\textbf{Source}: My sisters very pretty. \\
\textbf{Target}: My sisters \textcolor{red}{are} very pretty.

\noindent \textbf{1.1.3 Object Pronoun Missing}\\
\textbf{Definition:} Missing object pronouns (direct or indirect) in places where verbs require them.

\textbf{Source}: Donald is mean so no one likes. \\
\textbf{Target}: Donald is mean so no one likes \textcolor{red}{him}.

\noindent \textbf{1.1.4 Subject Pronoun Missing}\\
\textbf{Definition:} Missing subject pronouns where they are required, especially after subordinate conjunctions.

\textbf{Source}: He worked until fell over. \\
\textbf{Target}: He worked until \textcolor{red}{he} fell over.

\noindent \textbf{1.2 Misordered Parts}\\
\textbf{Definition:} Errors involving incorrect word order that disrupt the standard syntactic structure of English sentences.

\noindent \textbf{1.2.1 Verb Before Subject}\\
\textbf{Definition:} Incorrect word order where the verb precedes the subject in declarative sentences.

\textbf{Source}: \textcolor{red}{Escaped the professor} from prison. \\
\textbf{Target}: \textcolor{red}{The professor escaped} from prison.

\noindent \textbf{1.2.2 Subject and Object Permuted}\\
\textbf{Definition:} Errors occur when subject and object positions are reversed in declarative sentences.

\textbf{Source}: \textcolor{red}{English} use \textcolor{red}{many countries}. \\
\textbf{Target}: \textcolor{red}{Many countries} use \textcolor{red}{English}.

\noindent \textbf{2 The Auxiliary System}\\
\textbf{Definition:} Errors in the misuse, omission, or misalignment of auxiliary verbs ("do," "have," "be," modals) in questions, negatives, affirmatives, tense structures, or tag questions.

\noindent \textbf{2.1 The Use of \textit{DO}}\\
\textbf{Definition:} Errors related to the auxiliary verb "do," including its overuse, underuse, or misuse in forming questions, negatives, affirmatives, or maintaining consistent tense structures in English clauses.

\noindent \textbf{2.1.1 Overuse in Questions and Negatives}\\
\textbf{Definition:} Incorrect use of "do" in questions or negatives, where it appears with modal auxiliaries (can, must, should, etc.) or other auxiliaries (have, be) that already fulfill the grammatical function.

\textbf{Source}: \textcolor{red}{Does} she \textcolor{red}{have} come yet? \\
\textbf{Target}: \textcolor{red}{Has} she come yet?

\noindent \textbf{2.1.2 Underuse in Questions}\\
\textbf{Definition:} Failure to include do in questions when there are no auxiliary verbs, leading to incorrect word order or ungrammatical structures.

\textbf{Source}: Why we bow to each other? \\
\textbf{Target}: Why \textcolor{red}{do} we bow to each other?

\noindent \textbf{2.1.3 Overuse in Affirmative Sentences}\\
\textbf{Definition:} "Do" appears in a clause if there is no auxiliary, and the clause is a question or a negative. It does not appear in affirmative sentences.

\textbf{Source}: I \textcolor{red}{did leave} yesterday. \\
\textbf{Target}: I \textcolor{red}{left} yesterday.

\noindent \textbf{2.1.4 DO Missing from Negatives}\\
\textbf{Definition:} The auxiliary verb "do" is missing in negative sentences without other auxiliaries. In such cases, "do" must appear, and "not" must follow it to form the correct structure.

\textbf{Source}: I \textcolor{red}{practice not} religion. \\
\textbf{Target}: I \textcolor{red}{do not practice} religion.

\noindent \textbf{2.1.5 Tense Misplacement}\\
\textbf{Definition:} Errors involving placing tense on more than one verbal word or using conflicting tenses in a clause.

\textbf{Source}: \textcolor{red}{Do} you saw her already? \\
\textbf{Target}: \textcolor{red}{Did} you see her already?

\noindent \textbf{2.2 The Auxiliaries \textit{HAVE} and \textit{BE}}\\
\textbf{Definition:} Errors involving the misuse, omission, or incorrect combination of "have" and "be" in forming perfect, progressive, passive constructions, or linking non-verb predicates.

\noindent \textbf{2.2.1 Misformation of Perfect and Progressive Aspects}\\
\textbf{Definition:} Errors involving the incorrect formation of either the perfect aspect (combination of "have" and past participle) or the progressive aspect (combination of "be" and present participle).

\textbf{Source}: I have \textcolor{red}{saw} Broadway. \\
\textbf{Target}: I have \textcolor{red}{seen} Broadway.

\noindent \textbf{2.2.2 Passive Auxiliary Misformation}\\
\textbf{Definition:} Errors involving the omission or misuse of "be" in forming the passive voice.

\textbf{Source}: I \textcolor{red}{have} impressed with Plato. \\
\textbf{Target}: I \textcolor{red}{am} impressed with Plato.

\noindent \textbf{2.2.3 BE Missing}\\
\textbf{Definition:} Errors caused by omitting "be" before non-verb predicates (e.g., adjectives, nouns, adverbs, etc.), specifically in sentences that are not progressive or passive constructions.

\textbf{Source}: The bus always full of people. \\
\textbf{Target}: The bus \textcolor{red}{is} always full of people.

\noindent \textbf{2.2.4 DO Misused with BE}\\
\textbf{Definition:} Errors caused by incorrectly using "do" before "be" in questions or negatives.

\textbf{Source}: \textcolor{red}{Do} they \textcolor{red}{be} happy? \\
\textbf{Target}: \textcolor{red}{Are} they happy?

\noindent \textbf{2.3 Modals}\\
\textbf{Definition:} Errors in the use of modal verbs, affecting their grammatical role or the structure of the sentence.

\noindent \textbf{2.3.1 Misformation of the Next Verbal Word}\\
\textbf{Definition:} Errors involving the use of affixed forms (e.g., -ing, -ed) or the addition of "to" after modal verbs, which require a base form of the verb.

\textbf{Source}: I can \textcolor{red}{going} if you can. \\
\textbf{Target}: I can \textcolor{red}{go} if you can.

\noindent \textbf{2.3.2 Misunderstanding of Tense with Modals}\\
\textbf{Definition:} Errors stemming from a misunderstanding of modal tense properties.

\textbf{Source}: I must \textcolor{red}{can} catch this train. \\
\textbf{Target}: I must catch this train.

\noindent \textbf{2.4 Mismatching Auxiliaries in Tag Questions}\\
\textbf{Definition:} Errors caused by using a different auxiliary in the tag question than the one used in the main clause.

\textbf{Source}: She has been smoking less, \textcolor{red}{isn’t it}? \\
\textbf{Target}: She has been smoking less, \textcolor{red}{hasn’t she}?

\noindent \textbf{3 Passive Sentences}\\
\textbf{Definition:} Errors in passive voice construction or usage, including verb form issues, structural mismatches, preposition errors, or inappropriate application in certain contexts.

\noindent \textbf{3.1 Problems with Formation of Passive Sentences}\\
\textbf{Definition:} Errors in passive sentence construction, including misuse of "be," mismatched subject-verb relations, incorrect prepositions, or confusion between active and passive forms.

\noindent \textbf{3.1.1 Misformation of Passive Verb}\\
\textbf{Definition:} Errors caused by omitting or misplacing the auxiliary "be" in passive constructions, or failing to use the past participle form of the verb.

\textbf{Source}: The bread finished. \\
\textbf{Target}: The bread \textcolor{red}{is} finished.

\noindent \textbf{3.1.2 Active Order but Passive Form}\\
\textbf{Definition:} Errors caused by retaining the active sentence order while incorrectly using the passive form, resulting in a mismatch between the subject and the sentence structure.

\textbf{Source}: The \textcolor{red}{government was} forbidden \textcolor{red}{the people to grow opium}. \\
\textbf{Target}: The \textcolor{red}{people were} forbidden \textcolor{red}{to grow opium by the government}.

\noindent \textbf{3.1.3 Absent or Wrong Preposition Before Agent}\\
\textbf{Definition:} Errors caused by omitting or using the wrong preposition (e.g., "by") before the agent in a passive sentence.

\textbf{Source}: My brother was held up the traffic jam. \\
\textbf{Target}: My brother was held up \textcolor{red}{by} the traffic jam.

\noindent \textbf{3.1.4 Passive Order but Active Form}\\
\textbf{Definition:} Errors caused by using active verbs while following the word order of passive sentences, leading to mismatched sentence structure.

\textbf{Source}: English \textcolor{red}{use} many countries. \\
\textbf{Target}: English \textcolor{red}{is used} by many countries. \\

\noindent \textbf{3.2 Inappropriate Use of Passive}\\
\textbf{Definition:} Errors involving the misuse of passive voice, such as applying it to intransitive verbs or overextending it in complex sentences, leading to grammatical or logical inconsistencies.

\noindent \textbf{3.2.1 Making Intransitive Verbs Passive}\\
\textbf{Definition:} Errors caused by attempting to form the passive voice using intransitive verbs, which lack an object to become the subject in a passive construction.

\textbf{Source}: He \textcolor{red}{was arrived} early. \\
\textbf{Target}: He \textcolor{red}{arrived} early.

\noindent \textbf{3.2.2 Misusing Passives in Complex Sentences}\\
\textbf{Definition:} Errors caused by incorrectly applying passive constructions across multiple clauses in a complex sentence, leading to confusion about the subject and clause relations.

\textbf{Source}: \textcolor{red}{I was suggested by Mrs. Sena to} forget about this project. \\
\textbf{Target}: \textcolor{red}{Mrs. Sena suggested that I} forget about this project.

\noindent \textbf{4. Temporal Conjunctions}\\
\textbf{Definition:} Errors in the use, placement, or grammatical constructions associated with temporal conjunctions, including issues with clause structure, predicate types, tense agreement, and unnecessary future markers.

\noindent \textbf{4.1 Misplacement of Conjunctions}\\
\textbf{Definition:} Errors caused by incorrectly positioning conjunctions like "after," "since," or "while," leading to confusion about the sequence, causality, or timing of events in a sentence.

\noindent \textbf{4.1.1 Misplacement of After}\\
\textbf{Definition:} Errors involving incorrect placement or usage of "after" when describing the sequence of two events.

\textbf{Source}: I got up \textcolor{red}{after} I brushed my teeth. \\
\textbf{Target}: \textcolor{red}{After} I got up\textcolor{red}{,} I brushed my teeth.

\noindent \textbf{4.1.2 Misplacement of Since}\\
\textbf{Definition:} Errors involving incorrect placement or usage of "since" when describing the temporal relationship between two events.

\textbf{Source}: He broke his leg \textcolor{red}{since} he has thrown away his skis. \\
\textbf{Target}: \textcolor{red}{Since} he broke his leg\textcolor{red}{,} he has thrown away his skis.

\noindent \textbf{4.1.3 Misplacement of While}\\
\textbf{Definition:} Errors involving incorrect placement or usage of "while" to describe overlapping or interrupting events.

\textbf{Source}: \textcolor{red}{While} Getachew knocked on the door\textcolor{red}{,} I was doing my homework. \\
\textbf{Target}: Getachew knocked on the door \textcolor{red}{while} I was doing my homework.

\noindent \textbf{4.2 Form of Clauses After Temporal Conjunctions}\\
\textbf{Definition:} Errors in the structure of subordinate clauses following temporal conjunctions.

\noindent \textbf{4.2.1 Non-Finiteness of Subordinate Clauses}\\
\textbf{Definition:} Errors caused by omitting subjects or using incorrect verb forms in subordinate clauses, making them non-finite when they should be finite.

\textbf{Source}: After \textcolor{red}{him} goes, we will read a story. \\
\textbf{Target}: After \textcolor{red}{he} goes, we will read a story.

\noindent \textbf{4.2.2 Superfluous THAT}\\
\textbf{Definition:} Errors caused by inserting an unnecessary "that" immediately after a subordinate conjunction, disrupting sentence structure.

\textbf{Source}: Since \textcolor{red}{that} he has seen her, he has been cheerful. \\
\textbf{Target}: Since he has seen her, he has been cheerful.

\noindent \textbf{4.3 Selection of Predicate Types}\\
\textbf{Definition:} Errors in choosing appropriate predicate forms with temporal conjunctions.

\noindent \textbf{4.3.1 Confusion in Unlimited and Limited Verb Selection}\\
\textbf{Definition:} Errors caused by incorrect use of limited or unlimited verbs with temporal conjunctions such as "until" or "after".

\textbf{Source}: He \textcolor{red}{got} rich until he married. \\
\textbf{Target}: He \textcolor{red}{was} rich until he married.

\noindent \textbf{4.3.2 Difficulties in Changing Limitedness of Verbs}\\
\textbf{Definition:} Errors caused by using inherently limited verbs in constructions that require unlimited verbs.

\textbf{Source}: Since the child recovered from measles, he \textcolor{red}{grew} well. \\
\textbf{Target}: Since the child recovered from measles, he \textcolor{red}{has been growing} well.

\noindent \textbf{4.3.3 Misuse of Negatives with Temporal Conjunctions}\\
\textbf{Definition:} Errors involving the use of negatives with temporal conjunctions, where the clause requires a specific form of limited or unlimited verb.

\textbf{Source}: I did it while they \textcolor{red}{didn’t look}. \\
\textbf{Target}: I did it while they \textcolor{red}{weren't looking}.

\noindent \textbf{4.3.4 Misuse of End-of-the-Road Predicates}\\
\textbf{Definition:} Errors arising from the use of predicates that denote final or end states (e.g., dead, rotten, finished) in subordinate clauses with temporal conjunctions, where such predicates conflict with the expected progression or transition of states.

\textbf{Source}: The fruit had become rotten \textcolor{red}{until} we could eat it. \\
\textbf{Target}: The fruit had become rotten \textcolor{red}{before} we could eat it.

\noindent \textbf{4.4 Superficial Tense Agreement (STAGR)}\\
\textbf{Definition:} Errors in tense consistency between clauses linked by temporal conjunctions.

\noindent \textbf{4.4.1 Failure to Apply STAGR with BEFORE, AFTER, UNTIL, WHILE, WHEN}\\
\textbf{Definition:} Errors caused by mismatched tenses in clauses connected by temporal conjunctions. This typically occurs when the first verb in each clause does not agree in tense.

\textbf{Source}: When you were here yesterday you \textcolor{red}{have} promised to send your picture. \\
\textbf{Target}: When you were here yesterday you promised to send your picture.

\noindent \textbf{4.4.2 Inconsistency in Perfect Use: WHILE}\\
\textbf{Definition:} Errors occurring when one clause in a while construction uses the perfect aspect, but the other clause does not. Both clauses must either use the perfect aspect or omit it.

\textbf{Source}: While you \textcolor{red}{have worked}, I \textcolor{red}{make} phone \textcolor{red}{calls}. \\
\textbf{Target}: While you \textcolor{red}{worked}, I \textcolor{red}{made a} phone \textcolor{red}{call}.

\noindent \textbf{4.4.3 STAGR Misapplied: SINCE}\\
\textbf{Definition:} Errors caused by applying the STAGR rule (matching tenses) to since clauses, where since requires specific tense relationships between clauses.

\textbf{Source}: They \textcolor{red}{are} studying in this school since they are six years old. \\
\textbf{Target}: They \textcolor{red}{have been} studying in this school since they were six years old.

\noindent \textbf{4.5 Superfluous WILL and Other Future Constructions}\\
\textbf{Definition:} Errors involving unnecessary use of \textit{will} or other future constructions in subordinate clauses when the main clause already indicates future tense.

\textbf{Source}: We will eat after we \textcolor{red}{will} pray. \\
\textbf{Target}: We will eat after we pray.

\noindent \textbf{5. Sentential Complements}\\
\textbf{Definition:} Errors in subordinate clauses or complements, including word order, surrogate subjects, infinitives, gerunds, or complement forms required by the main verb.

\noindent \textbf{5.1 Misordering in Subordinate Constructions}\\
\textbf{Definition:} Errors caused by incorrect word order in subordinate constructions, where clauses deviate from the subject-verb-object (SVO) pattern or verbs are misplaced.

\textbf{Source}: He says that he \textcolor{red}{no money has}. \\
\textbf{Target}: He says that he \textcolor{red}{has no money}.

\noindent \textbf{5.2 Problems with Extraposition of Fat Subject}\\
\textbf{Definition:} Errors involving the omission or misuse of surrogate subjects like "it" or "there" when a heavy subject, such as a that-clause, is moved to the end of the sentence.

\noindent \textbf{5.2.1 Omission of Surrogate Subject}\\
\textbf{Definition:} Missing the word "it" as the subject when a heavy subject, such as a that-clause, is moved to the end of the sentence.

\textbf{Source}: \textcolor{red}{Is} very hard for me to learn English right. \\
\textbf{Target}: \textcolor{red}{It is} very hard for me to learn English right.

\noindent \textbf{5.2.2 Wrong Surrogate Subject: IT and THERE}\\
\textbf{Definition:} Errors caused by using incorrect surrogate subjects, such as he or that, instead of it or there, in sentences describing the weather, ambient conditions, or with extra posed subjects.

\textbf{Source}: \textcolor{red}{That} is funny to see him today. \\
\textbf{Target}: It is funny to see him today.

\noindent \textbf{5.3 Problems with Infinitives and Gerunds}\\
\textbf{Definition:} Errors in the use of infinitives and gerunds.

\noindent \textbf{5.3.1 Leaving Out the Subject}\\
\textbf{Definition:} Errors occur when the subject of a subordinate clause is omitted incorrectly, particularly when the subject is not repetitive or differs from the main clause subject.

\textbf{Source}: I think \textcolor{red}{to} have my I.D. card in here. \\
\textbf{Target}: I think \textcolor{red}{I} have my I.D. card in here.

\noindent \textbf{5.3.2 Misformations with Non-Nominative Subjects}\\
\textbf{Definition:} Errors occur when a non-nominative subject is incorrectly used with an infinitive or a gerund, instead of the correct form.

\textbf{Source}: \textcolor{red}{Him} to be so rich is unfair. \\
\textbf{Target}: \textcolor{red}{For him} to be so rich is unfair.

\noindent \textbf{5.3.3 Misformations Without Subjects}\\
\textbf{Definition:} Errors occur when "for" is mistakenly used before an infinitive that lacks a subject, instead of simply using "to" before the verb.

\textbf{Source}: It is necessary \textcolor{red}{for} finish the work. \\
\textbf{Target}: It is necessary \textcolor{red}{to} finish the work.

\noindent \textbf{5.3.4 Special Problems with MAKE, LET, HAVE, FIND}\\
\textbf{Definition:} Errors occur when using make, let, have, or find with infinitives or complements, such as unnecessarily adding "to" or omitting the required "it" as a surrogate subject.

\textbf{Source}: The vacuum cleaner makes easy to clean the house. \\
\textbf{Target}: The vacuum cleaner makes \textcolor{red}{it} easy to clean the house.

\noindent \textbf{5.3.5 Snatched Subject as Subject of Main Clause}\\
\textbf{Definition:} Errors occur when subject snatching is applied to adjectives or predicates that do not allow it, resulting in ungrammatical constructions.

\textbf{Source}: \textcolor{red}{The President is impossible to} be reelected. \\
\textbf{Target}: \textcolor{red}{It is impossible that the President will} be reelected.

\noindent \textbf{5.3.6 Snatched Subject as Object of Main Clause}\\
\textbf{Definition:} Errors occur when a verb that does not permit subject snatching is used incorrectly to create a construction where the subordinate subject becomes the main clause's object.

\textbf{Source}: \textcolor{red}{A girl} was decided \textcolor{red}{to} play the piano. \\
\textbf{Target}: \textcolor{red}{It} was decided \textcolor{red}{that} a girl \textcolor{red}{would} play the piano.

\noindent \textbf{5.3.7 Misformation of Gerunds After Prepositions}\\
\textbf{Definition:} Errors occur when the complement following a preposition is not in the required gerund or nominal form.

\textbf{Source}: We look forward to \textcolor{red}{see} you again. \\
\textbf{Target}: We look forward to \textcolor{red}{seeing} you again.

\noindent \textbf{5.4 Choosing Complement Types by Main Verb Meaning}\\
\textbf{Definition:} Errors in selecting the correct complement form (e.g., that-clauses, infinitives, or gerunds) based on the meaning and requirements of the main verb.

\noindent \textbf{5.4.1 Forms Taken by Propositions and Actions}\\
\textbf{Definition:} Errors occur when the incorrect complement form is used after verbs. Verbs like think or believe require a that-clause, while verbs like want or stop typically take gerunds or infinitives.

\textbf{Source}: Mark thinks the beans \textcolor{red}{needing} fertilizer. \\
\textbf{Target}: Mark thinks \textcolor{red}{that} the beans \textcolor{red}{need} fertilizer.

\noindent \textbf{5.4.2 Difficulty with Verbs Which Select Infinitives}\\
\textbf{Definition:} Errors occur when infinitives are required but incorrectly formed.

\textbf{Source}: I don’t expect \textcolor{red}{seeing} him today. \\
\textbf{Target}: I don’t expect \textcolor{red}{to see} him today.

\noindent \textbf{5.4.3 Difficulty with Verbs Which Select Gerunds}\\
\textbf{Definition:} Errors occur when gerunds are required but incorrectly formed.

\textbf{Source}: Don’t you remember \textcolor{red}{to} see her yesterday? \\
\textbf{Target}: Don't you remember \textcolor{red}{seeing} her yesterday?

\noindent \textbf{5.4.4 Confusion Over Complement Form After Auxiliaries}\\
\textbf{Definition:} Errors occur when the presence of auxiliaries leads to incorrect complement forms, as auxiliaries do not change the selection of complement forms determined by the main verb.

\textbf{Source}: I will enjoy \textcolor{red}{to swim}. \\
\textbf{Target}: I will enjoy \textcolor{red}{swimming}.

\noindent \textbf{6. Psychological Predicates}\\
\textbf{Definition:} Errors in constructing sentences with psychological predicates.

\noindent \textbf{6.1 Misordering of Subject and Object}\\
\textbf{Definition:} Errors caused by incorrect word order between subject and object, particularly with psychological verbs, leading to confusion in the roles of experiencer and stimulus.

\noindent \textbf{6.1.1 Misordering with Reverse Psychological Verbs}\\
\textbf{Definition:} Errors occur when the required stimulus-verb-experiencer word order for reverse psychological verbs is not followed, leading to confusion in meaning.

\textbf{Source}: The cat is on the dinner table, but \textcolor{red}{my father} doesn’t bother \textcolor{red}{that}. \\
\textbf{Target}: The cat is on the dinner table, but \textcolor{red}{that} doesn’t bother \textcolor{red}{my father}.

\noindent \textbf{6.1.2 Misordering with Straightforward Psychological Verbs}\\
\textbf{Definition:} Errors occur when experiencer and stimulus roles are reversed for straightforward psychological verbs, which require experiencer-verb-stimulus order.

\textbf{Source}: \textcolor{red}{The party} enjoyed \textcolor{red}{Aziz}. \\
\textbf{Target}: \textcolor{red}{Aziz} enjoyed \textcolor{red}{the party}.

\noindent \textbf{6.2 Embedded Sentences with Reverse Verbs}\\
\textbf{Definition:} Errors in using reverse psychological verbs, including misused forms, incorrect prepositions, missing elements, or improper handling of complex stimuli.

\noindent \textbf{6.2.1 Using the Experiencer as Subject}\\
\textbf{Definition:} Errors occur when the experiencer is used as the subject of reverse verbs without converting the verb to its participial form with "ED" and adding "be."

\textbf{Source}: I \textcolor{red}{surprise} that he likes it. \\
\textbf{Target}: I \textcolor{red}{am surprised} that he likes it.

\noindent \textbf{6.2.2 Wrong Use of Prepositions with ED Forms}\\
\textbf{Definition:} Errors occur when incorrect prepositions are used after reverse verbs in their ED form, as each verb typically requires specific prepositions.

\textbf{Source}: We were all bored \textcolor{red}{about} his teaching. \\
\textbf{Target}: We were all bored \textcolor{red}{by} his teaching.

\noindent \textbf{6.2.3 Confusing ED and ING Forms of Reverse Verbs}\\
\textbf{Definition:} Errors occur when ING forms (which describe stimuli) are used instead of ED forms (which describe experiencers), or vice versa.

\textbf{Source}: I was \textcolor{red}{surprising} that he came. \\
\textbf{Target}: I was \textcolor{red}{surprised} that he came.

\noindent \textbf{6.2.4 Leaving Out Stimulus or Experiencer}\\
\textbf{Definition:} Errors occur when reverse psychological verbs omit either the stimulus or the experiencer, leading to incomplete or unclear sentences.

\textbf{Source}: Don’t go to that movie. \textcolor{red}{It bores}. \\
\textbf{Target}: Don’t go to that movie. \textcolor{red}{It's boring}.

\noindent \textbf{6.2.5 Mismanaged Extraposition}\\
\textbf{Definition:} Errors occur when reverse verbs with complex or long stimuli are misstructured, often leading to awkward sentences. These can be corrected by reordering the stimulus or using an extraposition with "it."

\textbf{Source}: Everyone \textcolor{red}{delights} that you won the lottery. \\
\textbf{Target}: Everyone \textcolor{red}{is delighted} that you won the lottery.

\noindent \textbf{6.3 Straightforward Adjectives}\\
\textbf{Definition:} Errors involving the misuse of straightforward adjectives, including treating them as reverse adjectives or misusing them as verbs instead of linking to appropriate verbs.

\noindent \textbf{6.3.1 Misordering with Straightforward Adjectives}\\
\textbf{Definition:} Errors occur when straightforward adjectives are treated as reverse adjectives, with the stimulus used as the subject instead of the experiencer.

\textbf{Source}: It was impatient \textcolor{red}{to me} to find out my grade. \\
\textbf{Target}: I was impatient to find out my grade.

\noindent \textbf{6.3.2 Misuse of Adjectives as Verbs}\\
\textbf{Definition:} Errors occur when adjectives are misused as verbs instead of appearing after linking verbs like "be," "become," or others such as "feel" or "seem."

\textbf{Source}: Kwame \textcolor{red}{sorries} so much that your wife is sick. \\
\textbf{Target}: Kwame \textcolor{red}{is sorry} so much that your wife is sick.

\noindent \textbf{6.4 Reverse Adjectives}\\
\textbf{Definition:} Errors in the use of reverse adjectives, including misplacing the stimulus and experiencer, misordering in embedded sentences, or mishandling causation with impersonal adjectives.

\noindent \textbf{6.4.1 Misordering with Reverse Adjectives}\\
\textbf{Definition:} Errors occur when reverse adjectives are misused as straightforward adjectives, placing the experiencer as the subject instead of the stimulus.

\textbf{Source}: \textcolor{red}{I am hard} to get anything done. \\
\textbf{Target}: \textcolor{red}{It is hard for me} to get anything done.

\noindent \textbf{6.4.2 Misordering in Embedded Sentences}\\
\textbf{Definition:} Errors occur when reverse adjectives in embedded sentences place the experiencer as the subject instead of the stimulus.

\textbf{Source}: He admits \textcolor{red}{me hard} to learn quickly. \\
\textbf{Target}: He admits \textcolor{red}{(that) it is hard for me} to learn quickly.

\noindent \textbf{6.4.3 Difficulties with Causation}\\
\textbf{Definition:} Errors occur when causation is expressed incorrectly, often by misplacing objects or failing to use "it" with impersonal adjectives like "impossible" or "easy."

\textbf{Source}: The kids make impossible me to work. \\
\textbf{Target}: The kids make \textcolor{red}{it} impossible for me to work.

\end{tcolorbox}

\subsection{BRY17}
\label{subsec: errant}
\begin{figure*}[htbp]
    \centering
    \includegraphics[width=\textwidth]{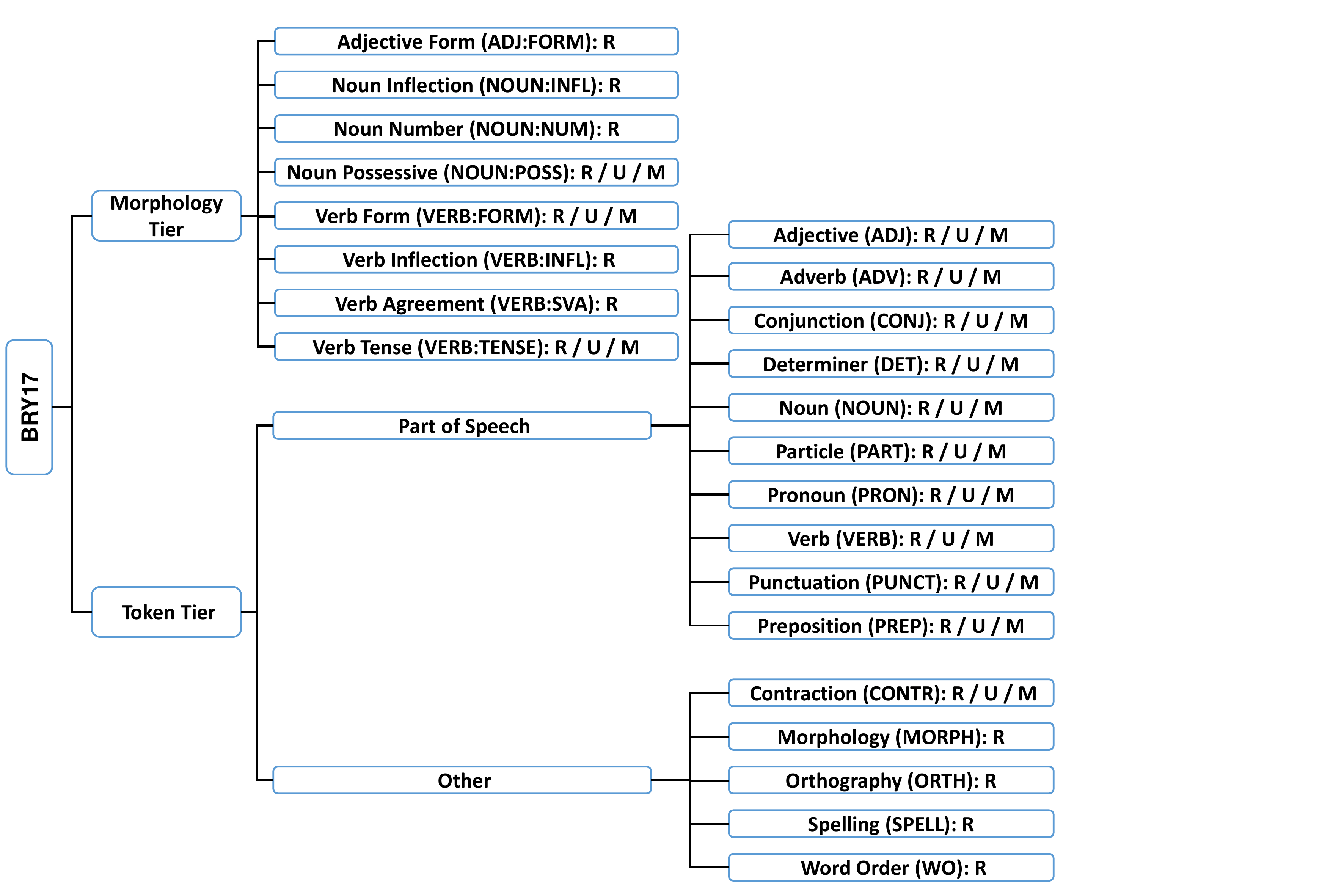}
    \caption{Overview of the BRY17 Error Classification Taxonomy}
    \label{fig:bry}
\end{figure*}

This error taxonomy is based on the Part-of-Speech (PoS) tags and the granularity level of the error.
Any error type in the taxonomy is prefixed with 'M:', 'R:', or 'U:', depending on whether it describes a \textbf{Missing}, \textbf{Replacement}, or \textbf{Unnecessary} edit respectively.
Each category contains subcategories with definitions and examples for clarity. Figure \ref{fig:bry} provides an overview of this taxonomy. Below, we provide a detailed description of this taxonomy:

\begin{tcolorbox}[breakable]
\noindent \textbf{1 Morphology Tier}\\
\textbf{Definition:} Errors related to morphological structures such as inflection, agreement, and verb forms.

\noindent \textbf{1.1 Adjective Form (ADJ:FORM) [R]}\\
\textbf{Definition:} Errors in the form of adjectives, such as incorrect comparative or superlative forms.

\textbf{Source}: This house is \textcolor{red}{more} bigger than the other. \\
\textbf{Target}: This house is bigger than the other.

\noindent \textbf{1.2 Noun Inflection (NOUN:INFL) [R]}\\
\textbf{Definition:} Count-mass noun errors.

\textbf{Source}: His \textcolor{red}{advices were} not helpful. \\
\textbf{Target}: His \textcolor{red}{advice was} not helpful.

\noindent \textbf{1.3 Noun Number (NOUN:NUM) [R]}\\
\textbf{Definition:} Errors in the singular or plural form of nouns.

\textbf{Source}: She has many \textcolor{red}{friend}. \\
\textbf{Target}: She has many \textcolor{red}{friends}.

\noindent \textbf{1.4 Noun Possessive (NOUN:POSS) [R/U/M]}\\
\textbf{Definition:} Errors in the possessive case, including incorrect or missing apostrophes, or unnecessary possessive markers.

\textbf{Source}: This is \textcolor{red}{Johns} book. \\
\textbf{Target}: This is \textcolor{red}{John's} book.

\noindent \textbf{1.5 Verb Form (VERB:FORM) [R/U/M]}\\
\textbf{Definition:} Errors in verb form, such as misuse of gerunds, infinitives, or participles.

\textbf{Source}: He improved the dish by \textcolor{red}{to cook} it longer. \\
\textbf{Target}: He improved the dish by \textcolor{red}{cooking} it longer.

\noindent \textbf{1.6 Verb Inflection (VERB:INFL) [R]}\\
\textbf{Definition:} Misapplication of tense morphology.

\textbf{Source}: She \textcolor{red}{getted} a new car yesterday. \\
\textbf{Target}: She \textcolor{red}{got} a new car yesterday.

\noindent \textbf{1.7 Verb Agreement (VERB:SVA) [R]}\\
\textbf{Definition:} Subject-verb agreement errors, such as mismatched number or person.

\textbf{Source}: They \textcolor{red}{was} happy. \\
\textbf{Target}: They \textcolor{red}{were} happy.

\noindent \textbf{1.8 Verb Tense (VERB:TENSE) [R/U/M]}\\
\textbf{Definition:} Wrong choice of inflectional and periphrastic tense, modal verbs, and passivization.

\textbf{Source}: He \textcolor{red}{has} went to the park yesterday. \\
\textbf{Target}: He went to the park yesterday.

\noindent \textbf{2 Token Tier}\\
\textbf{Definition:} Errors related to individual tokens, including words and punctuation.

\noindent \textbf{2.1 Part of Speech}

\noindent \textbf{2.1.1 Adjective (ADJ) [R/U/M]}\\
\textbf{Definition:} Errors in the use, omission, or addition of adjectives.

\textbf{Source}: She is a \textcolor{red}{beauty} dancer. \\
\textbf{Target}: She is a \textcolor{red}{beautiful} dancer.

\noindent \textbf{2.1.2 Adverb (ADV) [R/U/M]}\\
\textbf{Definition:} Errors in the use, omission, or addition of adverbs.

\textbf{Source}: She sings \textcolor{red}{good}. \\
\textbf{Target}: She sings \textcolor{red}{well}.

\noindent \textbf{2.1.3 Conjunction (CONJ) [R/U/M]}\\
\textbf{Definition:} Errors in the use, omission, or addition of conjunctions.

\textbf{Source}: She likes apples oranges. \\
\textbf{Target}: She likes apples \textcolor{red}{and} oranges.

\noindent \textbf{2.1.4 Determiner (DET) [R/U/M]}\\
\textbf{Definition:} Errors in the use, omission, or addition of determiners.

\textbf{Source}: She bought the \textcolor{red}{a} apple. \\
\textbf{Target}: She bought the apple.

\noindent \textbf{2.1.5 Noun (NOUN) [R/U/M]}\\
\textbf{Definition:} Errors in the use, omission, or addition of nouns.

\textbf{Source}: She brought her friend \textcolor{red}{friend}. \\
\textbf{Target}: She brought her friend.

\noindent \textbf{2.1.6 Particle (PART) [R/U/M]}\\
\textbf{Definition:} Errors involving particles, such as their incorrect placement, omission, or addition.

\textbf{Source}: He got \textcolor{red}{up} into the car. \\
\textbf{Target}: He got into the car.

\noindent \textbf{2.1.7 Pronoun (PRON) [R/U/M]}\\
\textbf{Definition:} Errors involving pronouns, such as incorrect form, extra pronouns, or omitted pronouns.

\textbf{Source}: She \textcolor{red}{herself} went to the store herself. \\
\textbf{Target}: She went to the store herself.

\noindent \textbf{2.1.8 Verb (VERB) [R/U/M]}\\
\textbf{Definition:} Errors involving the incorrect verb form, addition of redundant verbs, or omission of verbs.

\textbf{Source}: He to school every day. \\
\textbf{Target}: He \textcolor{red}{goes} to school every day.

\noindent \textbf{2.1.9 Punctuation (PUNCT) [R/U/M]}\\
\textbf{Definition:} Errors in punctuation, including incorrect use, extra punctuation marks, or missing punctuation.

\textbf{Source}: She said, "Hello\textcolor{red}{".} \\
\textbf{Target}: She said, "Hello\textcolor{red}{."}

\noindent \textbf{2.1.10 Preposition (PREP) [R/U/M]}\\
\textbf{Definition:} Errors in the use, omission, or addition of prepositions.

\textbf{Source}: He is good math. \\
\textbf{Target}: He is good \textcolor{red}{at} math.

\noindent \textbf{2.2 Other}

\noindent \textbf{2.2.1 Contraction (CONTR) [R/U/M]}\\
\textbf{Definition:} Errors in the use, omission, or addition of contractions.

\textbf{Source}: I \textcolor{red}{isn't} ready for the exam. \\
\textbf{Target}: I\textcolor{red}{'m not} ready for the exam.

\noindent \textbf{2.2.2 Morphology (MORPH) [R]}\\
\textbf{Definition:} Errors where tokens share the same lemma but differ in other grammatical or syntactical attributes.

\textbf{Source}: The boy \textcolor{red}{runned} fast. \\
\textbf{Target}: The boy \textcolor{red}{ran} fast.

\noindent \textbf{2.2.3 Orthography (ORTH) [R]}\\
\textbf{Definition:} Errors in capitalization, whitespace, or other orthographic conventions.

\textbf{Source}: \textcolor{red}{he} went to School. \\
\textbf{Target}: \textcolor{red}{He} went to school.

\noindent \textbf{2.2.4 Spelling (SPELL) [R]}\\
\textbf{Definition:} Errors in the spelling of words.

\textbf{Source}: She \textcolor{red}{recieved} the package. \\
\textbf{Target}: She \textcolor{red}{received} the package.

\noindent \textbf{2.2.5 Word Order (WO) [R]}\\
\textbf{Definition:} Errors in the arrangement of words within a sentence.

\textbf{Source}: \textcolor{red}{To school she goes} every day. \\
\textbf{Target}: \textcolor{red}{She goes to school} every day.

\end{tcolorbox}

\subsection{FEI23}
\label{subsec: enhancing}
This taxonomy classifies grammatical errors into three levels based on cognitive complexity (\textbf{Single-word level}, \textbf{Inter-word level}, and \textbf{Discourse-level}) addressing different aspects of language learning and comprehension. Each category contains subcategories with definitions and examples for clarity. Figure \ref{fig:fei} provides an overview of this taxonomy. Below, we provide a detailed description of this taxonomy:

\begin{figure*}[tb]
    \centering
    \includegraphics[width=\textwidth, trim=0 0 230 80, clip]{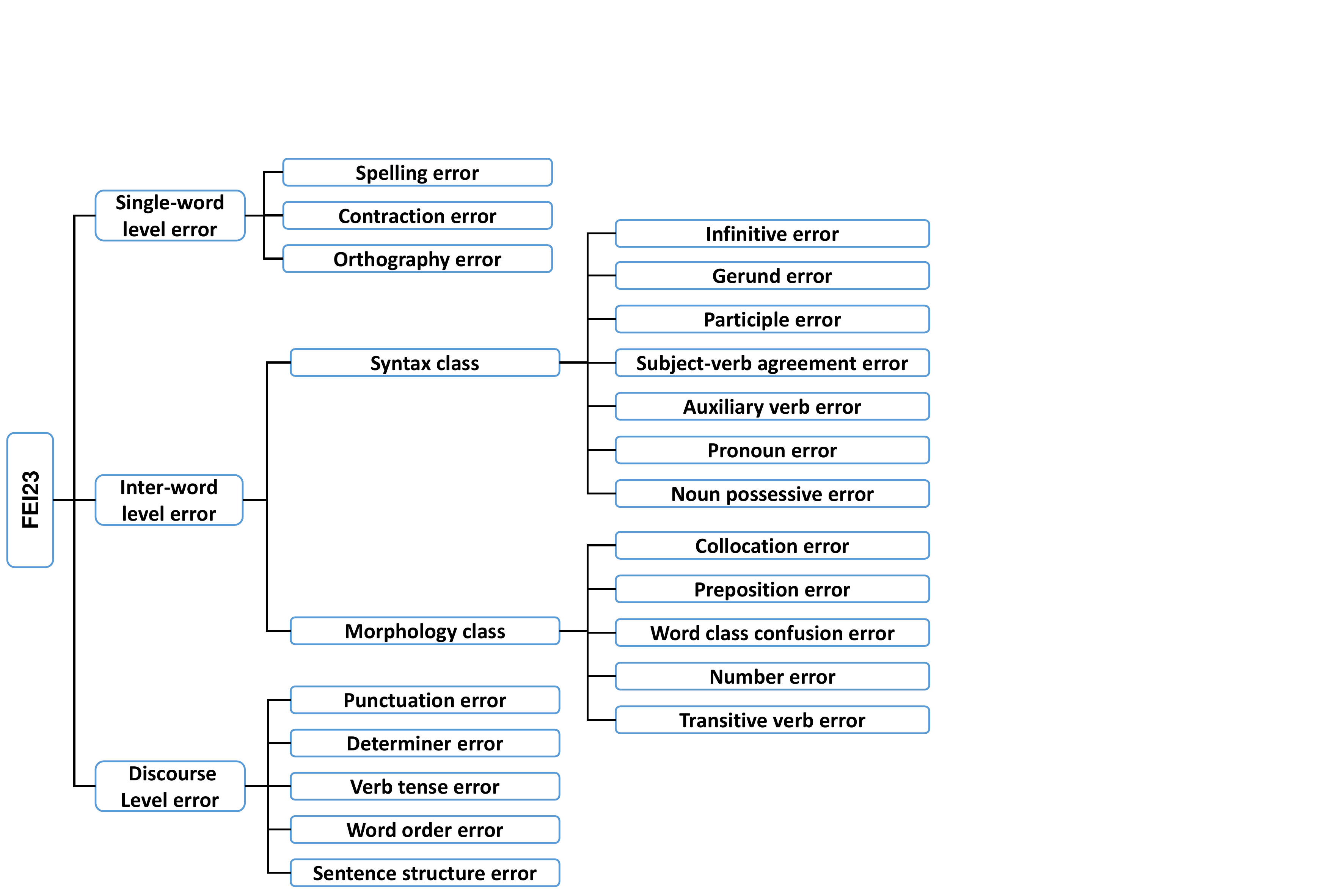}
    \caption{Overview of the FEI23 Error Classification Taxonomy}
    \label{fig:fei}
\end{figure*}

\begin{tcolorbox}[breakable]
\noindent \textbf{1 Single-word Level Error}\\
\textbf{Definition:} Errors confined to a single word, typically involving spelling, contractions, or capitalization.

\noindent \textbf{1.1 Spelling Error}\\
\textbf{Definition:} Incorrect spelling of a word.

\textbf{Source}: She \textcolor{red}{recieved} the letter. \\
\textbf{Target}: She \textcolor{red}{received} the letter.

\noindent \textbf{1.2 Contraction Error}\\
\textbf{Definition:} Incorrect use or formation of contractions.

\textbf{Source}: I \textcolor{red}{isn't} ready. \\
\textbf{Target}: I\textcolor{red}{'m not} ready.

\noindent \textbf{1.3 Orthography Error}\\
\textbf{Definition:} Errors in capitalization or other writing conventions.

\textbf{Source}: \textcolor{red}{he} went to school. \\
\textbf{Target}: \textcolor{red}{He} went to school.

\noindent \textbf{2 Inter-word Level Error}\\
\textbf{Definition:} Errors involving relationships between multiple words, including grammar, morphology, and word usage.

\noindent \textbf{2.1 Syntax Class}\\
\textbf{Definition:} Errors in sentence grammar and structure.

\noindent \textbf{2.1.1 Infinitive Error}\\
\textbf{Definition:} Errors like missing "to" before a verb in to-infinitives, or unnecessary "to" after modal verbs for zero-infinitives.

\textbf{Source}: I would like \textcolor{red}{going} home. \\
\textbf{Target}: I would like \textcolor{red}{to go} home.

\noindent \textbf{2.1.2 Gerund Error}\\
\textbf{Definition:} Misuse of the verb form that should act as a noun in a sentence.

\textbf{Source}: I enjoy \textcolor{red}{to play}. \\
\textbf{Target}: I enjoy \textcolor{red}{playing}.

\noindent \textbf{2.1.3 Participle Error}\\
\textbf{Definition:} Confusion between participles and ordinary verb tenses.

\textbf{Source}: She has \textcolor{red}{did} her homework. \\
\textbf{Target}: She has \textcolor{red}{done} her homework.

\noindent \textbf{2.1.4 Subject-verb Agreement Error}\\
\textbf{Definition:} The verb does not agree with the subject in number.

\textbf{Source}: They \textcolor{red}{was} happy. \\
\textbf{Target}: They \textcolor{red}{were} happy.

\noindent \textbf{2.1.5 Auxiliary Verb Error}\\
\textbf{Definition:} Misuse of auxiliary verbs such as "do," "have," or modal auxiliaries like "could," "may," "should."

\textbf{Source}: She can \textcolor{red}{sings} well. \\
\textbf{Target}: She can \textcolor{red}{sing} well.

\noindent \textbf{2.1.6 Pronoun Error}\\
\textbf{Definition:} Pronouns do not agree in number, person, or gender with their antecedents.

\textbf{Source}: \textcolor{red}{Her} went to the store. \\
\textbf{Target}: \textcolor{red}{She} went to the store.

\noindent \textbf{2.1.7 Noun Possessive Error}\\
\textbf{Definition:} Misuse of possessive adjectives and possessive nouns.

\textbf{Source}: This is \textcolor{red}{Johns} book. \\
\textbf{Target}: This is \textcolor{red}{John's} book.

\noindent \textbf{2.2 Morphology Class}\\
\textbf{Definition:} Errors in the structure or form of words, including their grammatical relationships.

\noindent \textbf{2.2.1 Collocation Error}\\
\textbf{Definition:} Atypical word combinations that are grammatically acceptable but not common.

\textbf{Source}: He \textcolor{red}{did} a crime. \\
\textbf{Target}: He \textcolor{red}{committed} a crime.

\noindent \textbf{2.2.2 Preposition Error}\\
\textbf{Definition:} Misuse of prepositional words.

\textbf{Source}: He is good \textcolor{red}{in} math. \\
\textbf{Target}: He is good \textcolor{red}{at} math.

\noindent \textbf{2.2.3 Word Class Confusion Error}\\
\textbf{Definition:} Confusions in part of speech, such as noun/adjective or adjective/adverb confusion.

\textbf{Source}: She is a \textcolor{red}{beauty} dancer. \\
\textbf{Target}: She is a \textcolor{red}{beautiful} dancer.

\noindent \textbf{2.2.4 Number Error}\\
\textbf{Definition:} Confusion in singular or plural form of nouns.

\textbf{Source}: She has many \textcolor{red}{friend}. \\
\textbf{Target}: She has many \textcolor{red}{friends}.

\noindent \textbf{2.2.5 Transitive Verb Error}\\
\textbf{Definition:} Extra preposition after transitive verbs or missing preposition after intransitive verbs.

\textbf{Source}: He gave. \\
\textbf{Target}: He gave \textcolor{red}{a gift}.

\noindent \textbf{3 Discourse Level Error}\\
\textbf{Definition:} Errors affecting the overall structure, flow, or coherence of a sentence or discourse.

\noindent \textbf{3.1 Punctuation Error}\\
\textbf{Definition:} Errors in the use, omission, or addition of punctuation marks.

\textbf{Source}: She said, "Hello\textcolor{red}{".} \\
\textbf{Target}: She said, "Hello\textcolor{red}{."}

\noindent \textbf{3.2 Determiner Error}\\
\textbf{Definition:} Errors in the use, omission, or addition of determiners.

\textbf{Source}: She bought apple. \\
\textbf{Target}: She bought \textcolor{red}{an} apple.

\noindent \textbf{3.3 Verb Tense Error}\\
\textbf{Definition:} Incongruities in verb tenses, such as an erroneous tense shift in a compound sentence.

\textbf{Source}: He \textcolor{red}{go} to school yesterday. \\
\textbf{Target}: He \textcolor{red}{went} to school yesterday.

\noindent \textbf{3.4 Word Order Error}\\
\textbf{Definition:} Errors in arranging words in the correct sequence within a sentence.

\textbf{Source}: \textcolor{red}{To school she goes} every day. \\
\textbf{Target}: \textcolor{red}{She goes to school} every day.

\noindent \textbf{3.5 Sentence Structure Error}\\
\textbf{Definition:} Errors affecting the overall grammatical structure of a sentence.

\textbf{Source}: When he comes, I will leave \textcolor{red}{before he arrives}. \\
\textbf{Target}: When he comes, I will leave.

\end{tcolorbox}

\section{Taxonomy-Specific Issues Affecting Rationality Metrics}  
\label{appendix main}  
This appendix provides a detailed analysis of taxonomy-specific issues that impact the rationality metrics assessed in our study.  

POL73 contains categories such as \textit{indefinite article incorrect} and \textit{determiners}, which often capture the same errors, leading to overlapping classifications. Similarly, \textit{inappropriate words but correct word class} and \textit{general semantic confusion} lack clear differentiation, reducing mutual exclusivity. TUC74 introduces ambiguous categories, such as \textit{simple predicate missing: be} vs. \textit{be missing}, and \textit{surrogate subject missing: there/it} vs. \textit{omission of surrogate subject}, creating annotation inconsistencies. FEI23, by structuring categories based on cognitive levels, ensures high exclusivity by clearly defining error boundaries and reducing ambiguity.  

POL73 includes categories like \textit{new creations} but lacks fundamental categories such as \textit{punctuation error} and \textit{orthography error}, slightly reducing its coverage. TUC74 performs worse in coverage as it omits not only \textit{punctuation error} and \textit{orthography error} but also \textit{spelling error}. This forces annotators to rely excessively on the \textit{other} category, significantly lowering its coverage score. BRY17, by defining error types based on both part-of-speech and token-level operations, ensures a broader representation of errors, leading to the highest coverage among all taxonomies.  

TUC74 demonstrates extreme specificity by defining five separate categories for errors related to temporal conjunctions, such as \textit{misplacement of while}. This level of granularity results in sparsely populated error categories, skewing distributional balance. POL73 takes the opposite approach, overgeneralizing error types. The category \textit{inappropriate words and incorrect word class} encompasses a wide range of unrelated grammatical issues, resulting in an overly broad classification that reduces distinction between different error types. FEI23 adopts a hierarchical classification strategy at three levels—\textit{single-word level}, \textit{inter-word level}, and \textit{discourse level}—ensuring balanced categorization without excessive fragmentation or aggregation.  

POL73 and TUC74 include overly technical or abstract error categories, such as \textit{phonetic similarity within English} and \textit{misuse of end-of-the-road predicates}, making them difficult to interpret and apply consistently. BRY17 and FEI23, by contrast, use straightforward category labels with simple lexical definitions, improving both human annotation efficiency and model interpretability. Their classification schemes facilitate direct mapping to computational models, reducing the complexity of automated grammar error detection.  

These taxonomy-specific issues highlight the challenges in designing effective error classification taxonomies and their impact on the rationality metrics assessed in our study.  

\section{Ablation Study on Category Fusion}
\label{appendix ablation study}
To examine the impact of classification granularity on four metrics, we conduct an ablation study by merging specific error categories. The specific taxonomy structures before and after fusion are detailed in the Appendix \ref{appendix ablation}. To ensure control for variability in model predictions, all results in this study are only derived from ChatGPT-4o. The experimental results are shown in Table \ref{table ablation}.

\begin{table*}[htb]
\centering
\resizebox{\textwidth}{!}{
\begin{tabular}{l c c c c}
\toprule
\textbf{Taxonomies} & \textbf{Exclusivity ($\tau =0.7)$ $\uparrow$} & \textbf{Coverage $\uparrow$} & \textbf{Balance $\uparrow$} & \textbf{Usability (Macro F1 / Micro F1) $\uparrow$} \\
\midrule
POL73 & 0.842 & 0.698 & \textbf{0.687} & 0.301 / 0.478 \\
POL73 (Fusion) & \textbf{0.886} & \textbf{0.715} & 0.662 & \textbf{0.371} / \textbf{0.524} \\
\midrule
TUC74 & 0.703 & 0.160 & \textbf{0.210} & 0.061 / 0.099 \\
TUC74 (Fusion) & \textbf{0.774} & \textbf{0.183} & 0.178 & \textbf{0.131} / \textbf{0.113} \\
\midrule
BRY17 & 0.921 & 0.979 & \textbf{0.829} & 0.610 / 0.760 \\
BRY17 (Fusion) & \textbf{0.932} & \textbf{0.984} & 0.786 &\textbf{ 0.618 }/ \textbf{0.800 }  \\
\midrule
FEI23 & \textbf{0.877} & 0.924 & \textbf{0.878} & \textbf{0.631} / \textbf{0.743} \\
FEI23 (Fusion) & 0.830 & \textbf{0.925} & 0.834 & 0.616 / 0.719 \\
\bottomrule
\end{tabular}}
\caption{Ablation Study on the Impact of Merging Error Categories (ChatGPT-4o). Taxonomies labeled with Fusion indicate error classification taxonomies after category merging.}
\label{table ablation}
\end{table*}

Our findings indicate that the fusion of error categories leads to a consistent increase in Coverage across all classification taxonomies. This is expected, as the newly formed parent category inherently encompasses a broader scope than its individual subcategories. By expanding the definition of error categories, previously unclassified errors are now incorporated into the taxonomy, leading to a higher coverage score.

Conversely, Balance decreases across all taxonomies after category fusion. This decline can be attributed to the inherent structure of the original taxonomies: error categories with broad definitions already contained a substantial number of samples prior to fusion. Merging smaller categories into these broader ones exacerbates the data imbalance, further skewing the error distribution. As a result, the overall balance of the taxonomy deteriorates.

The impact on Exclusivity varies depending on the specific taxonomy and the relationships between merged categories. For instance, in BRY17, the pre-fusion taxonomy contained overlapping error categories such as \textit{particle} and \textit{preposition}. By merging these, the overlap is eliminated, leading to an increase in exclusivity. However, in FEI23, pre-fusion error types such as \textit{contraction error}, \textit{spelling error}, and \textit{orthography error} were mutually exclusive and did not overlap with other categories. After merging them into the broader \textit{single-word level error}, overlaps emerged with other categories like \textit{pronoun error}, resulting in a decrease in exclusivity.

Similarly, Usability is influenced by classification granularity, as broader error categories reduce the model's ability to distinguish fine-grained errors. Merging categories weakens decision boundaries, increasing intra-category variability and making classification more ambiguous. While higher Exclusivity can enhance usability by reducing overlaps, it is not the sole factor—when fusion introduces ambiguity, usability declines due to the model's reduced classification accuracy. Our results confirm that category fusion affects usability by altering decision boundaries and classification effectiveness.

Our ablation study on category fusion demonstrates that increasing classification granularity leads to higher Coverage by broadening error definitions, but decreases Balance due to exacerbated data imbalance. The effect on Exclusivity depends on pre-fusion category overlap, improving when redundant categories are merged but declining when new overlaps emerge. The impact on Usability varies depending on how category fusion reshapes classification boundaries—some fusions improve clarity, while others introduce ambiguity that weakens classification effectiveness. Our findings highlight an important challenge in error taxonomy design: both excessive simplification and over-fragmentation can negatively impact classification taxonomies. Constructing an effective taxonomy is highly complex and cannot be achieved solely through empirical intuition, underscoring the importance of rigorous evaluation metrics in assessing classification taxonomies.

\section{Error Category Fusion Details}
\label{appendix ablation}

\subsection{POL73}

Before Fusion:
\begin{tcolorbox}[breakable]
\noindent \textbf{2.2 Verb Phrase}\\
\textbf{Definition:} Errors related to verb use, including tense, form, and omission.

\noindent \textbf{2.2.1 Omission of Verb}\\
\textbf{Definition:} Errors involving the omission of a necessary verb, including main verbs and auxiliary verbs such as forms of "be".

\textbf{Source}: He to school yesterday. \\
\textbf{Target}: He \textcolor{red}{went} to school yesterday.

\noindent \textbf{2.2.2 Use of Progressive Tense}\\
\textbf{Definition:} Errors in forming or using the progressive tense, including omission of the auxiliary verb "be", incorrect use of the "-ing" form, or substitution of the progressive where it is not expected.

\textbf{Source}: He is \textcolor{red}{play} football. \\
\textbf{Target}: He is \textcolor{red}{playing} football.

\noindent \textbf{2.2.3 Agreement of Subject and Verb}\\
\textbf{Definition:} Errors where the subject and verb fail to agree in number, person, or tense.

\textbf{Source}: I didn't know what it \textcolor{red}{is}. \\
\textbf{Target}: I didn't know what it \textcolor{red}{was}.

\end{tcolorbox}
\noindent After Fusion:
\begin{tcolorbox}[breakable]
\noindent \textbf{2.2 Verb Phrase}\\
\textbf{Definition:} Errors related to verb use, including tense, form, and omission.

\textbf{Source}: He to school yesterday. \\
\textbf{Target}: He \textcolor{red}{went} to school yesterday.

\end{tcolorbox}

\subsection{TUC74}

Before Fusion:
\begin{tcolorbox}[breakable]
\noindent \textbf{1.1 Missing Parts}\\
\textbf{Definition:} Errors involving the omission of essential sentence components necessary for grammatical completeness and structural clarity.

\noindent \textbf{1.1.1 Surrogate Subject Missing: there and it}\\
\textbf{Definition:} Missing placeholder subjects "there" or "it," which are required in English sentence structure.

\textbf{Source}: \textcolor{red}{Was} a riot last night. \\
\textbf{Target}: \textcolor{red}{There was} a riot last night.

\noindent \textbf{1.1.2 Simple Predicate Missing: \textit{be}}\\
\textbf{Definition:} Missing the simple predicate "be," which is required when the predicate consists of adjectives or noun phrases.

\textbf{Source}: My sisters very pretty. \\
\textbf{Target}: My sisters \textcolor{red}{are} very pretty.

\noindent \textbf{1.1.3 Object Pronoun Missing}\\
\textbf{Definition:} Missing object pronouns (direct or indirect) in places where verbs require them.

\textbf{Source}: Donald is mean so no one likes. \\
\textbf{Target}: Donald is mean so no one likes \textcolor{red}{him}.

\noindent \textbf{1.1.4 Subject Pronoun Missing}\\
\textbf{Definition:} Missing subject pronouns where they are required, especially after subordinate conjunctions.

\textbf{Source}: He worked until fell over. \\
\textbf{Target}: He worked until \textcolor{red}{he} fell over.

\end{tcolorbox}
\noindent After Fusion:
\begin{tcolorbox}[breakable]
\noindent \textbf{1.1 Missing Parts}\\
\textbf{Definition:} Errors involving the omission of essential sentence components necessary for grammatical completeness and structural clarity.

\textbf{Source}: My sisters very pretty. \\
\textbf{Target}: My sisters \textcolor{red}{are} very pretty.

\end{tcolorbox}

\subsection{BRY17}
Before Fusion:
\begin{tcolorbox}[breakable]
\noindent \textbf{2.1.3 Conjunction (CONJ) [R/U/M]}\\
\textbf{Definition:} Errors in the use, omission, or addition of conjunctions.

\textbf{Source}: She likes apples oranges. \\
\textbf{Target}: She likes apples \textcolor{red}{and} oranges.

\noindent \textbf{2.1.6 Particle (PART) [R/U/M]}\\
\textbf{Definition:} Errors involving particles, such as their incorrect placement, omission, or addition.

\textbf{Source}: He got \textcolor{red}{up} into the car. \\
\textbf{Target}: He got into the car.

\noindent \textbf{2.1.10 Preposition (PREP) [R/U/M]}\\
\textbf{Definition:} Errors in the use, omission, or addition of prepositions.

\textbf{Source}: He is good math. \\
\textbf{Target}: He is good \textcolor{red}{at} math.

\end{tcolorbox}
\noindent After Fusion:
\begin{tcolorbox}[breakable]
\noindent \textbf{2.1.8 Function Word (FUNC:WORD) [R/U/M]}\\
\textbf{Definition:} Errors in the use, omission, or addition of function words. These errors affect sentence structure and meaning by altering how words and phrases connect.

\textbf{Source}: She likes apples oranges. \\
\textbf{Target}: She likes apples \textcolor{red}{and} oranges.

\end{tcolorbox}

\subsection{FEI23}

Before Fusion:
\begin{tcolorbox}[breakable]
\noindent \textbf{1 Single-word Level Error}\\
\textbf{Definition:} Errors confined to a single word, typically involving spelling, contractions, or capitalization.

\noindent \textbf{1.1 Spelling Error}\\
\textbf{Definition:} Incorrect spelling of a word.

\textbf{Source}: She \textcolor{red}{recieved} the letter. \\
\textbf{Target}: She \textcolor{red}{received} the letter.

\noindent \textbf{1.2 Contraction Error}\\
\textbf{Definition:} Incorrect use or formation of contractions.

\textbf{Source}: I \textcolor{red}{isn't} ready. \\
\textbf{Target}: I\textcolor{red}{'m not} ready.

\noindent \textbf{1.3 Orthography Error}\\
\textbf{Definition:} Errors in capitalization or other writing conventions.

\textbf{Source}: \textcolor{red}{he} went to school. \\
\textbf{Target}: \textcolor{red}{He} went to school.

\end{tcolorbox}
\noindent After Fusion:
\begin{tcolorbox}[breakable]
\noindent \textbf{1 Single-word Level Error}\\
\textbf{Definition:} Errors confined to a single word, typically involving spelling, contractions, or capitalization.

\textbf{Source}: She \textcolor{red}{recieved} the letter. \\
\textbf{Target}: She \textcolor{red}{received} the letter.

\end{tcolorbox}

\end{document}